\documentclass[11pt]{article}

\usepackage[final]{acl}

\usepackage{amsmath}    
\usepackage{hyperref}
\usepackage{times}
\usepackage{latexsym}
\usepackage{xcolor}
\usepackage{pifont}
\usepackage{fontawesome}
\usepackage{enumitem}
\usepackage{amssymb}  
\usepackage{amsthm}  
\usepackage{bbm}
\usepackage{tcolorbox}
\usepackage{multirow}
\usepackage{booktabs}
\usepackage{arydshln}
\usepackage{colortbl}
\usepackage{makecell}  

\newcommand{\bluecheck}{\textcolor{blue}{\ding{52}}}
\newcommand{\greencheck}{\textcolor{green}{\ding{52}}}
\newcommand{\bluecross}{\textcolor{blue}{\ding{56}}}
\newcommand{\greencross}{\textcolor{green}{\ding{56}}}
\newcommand{\faCircleOMath}{\mbox{\faCircleO}}
\newcommand{\faCheckMath}{\mbox{\faCheck}}
\newcommand{\faCloseMath}{\mbox{\faClose}}
\newcommand{\bluecheckMath}{\mbox{\textcolor{blue}{\ding{52}}}}
\newcommand{\greencheckMath}{\mbox{\textcolor{green}{\ding{52}}}}
\newcommand{\bluecrossMath}{\mbox{\textcolor{blue}{\ding{56}}}}
\newcommand{\greencrossMath}{\mbox{\textcolor{green}{\ding{56}}}}

\usepackage{xspace}
\newcommand{\ourmethod}{{\fontfamily{lmtt}\selectfont \textbf{DTA}}\xspace}
\newcommand{\KBrag}{\mathrm{KB}_{\mathrm{rag}}}
\newcommand{\KBparam}{\mathrm{KB}_{\mathrm{param}}}
\newcommand{\KBr}{\mathrm{KB}_r}

\usepackage[T1]{fontenc}

\usepackage[utf8]{inputenc}

\usepackage{microtype}

\usepackage{inconsolata}

\usepackage{graphicx}

\title{Divide-Then-Align: Honest Alignment based on the Knowledge Boundary of RAG}

\author{
  \textbf{Xin Sun\textsuperscript{1,2}}\thanks{Equal contribution.},
  \textbf{Jianan Xie\textsuperscript{3}}\footnotemark[1],
  \textbf{Zhongqi Chen\textsuperscript{4}},
  \textbf{Qiang Liu\textsuperscript{2}}\thanks{Corresponding authors},
  \textbf{Shu Wu\textsuperscript{2}},
\\
  \textbf{Yuehe Chen\textsuperscript{4}},
  \textbf{Bowen Song\textsuperscript{4}}\footnotemark[2],
  \textbf{Weiqiang Wang\textsuperscript{4}}
  \textbf{Zilei Wang\textsuperscript{1}}
  \textbf{Liang Wang\textsuperscript{2}},
\\
  \textsuperscript{1}USTC \textsuperscript{2}NLPR, MAIS, CASIA \textsuperscript{3}SUSTech \textsuperscript{4}Independent \\
  \texttt{sunxin000@mail.ustc.edu.cn, 12110714@mail.sustech.edu.cn} \\
  \texttt{\{qiang.liu, shu.wu, wangliang\}@nlpr.ia.ac.cn, zlwang@ustc.edu.cn} \\
  \texttt{ \{chenzhongqi1997, a881465844, wdboou, wang.weiqiang\}@gmail.com}
}

\begin{document}
\maketitle
\begin{abstract}

Large language models (LLMs) augmented with retrieval systems have significantly advanced natural language processing tasks by integrating external knowledge sources, enabling more accurate and contextually rich responses. To improve the robustness of such systems against noisy retrievals, Retrieval-Augmented Fine-Tuning (RAFT) has emerged as a widely adopted method. However, RAFT conditions models to generate answers even in the absence of reliable knowledge. This behavior undermines their reliability in high-stakes domains, where acknowledging uncertainty is critical. To address this issue, we propose Divide-Then-Align (\ourmethod), a post-training approach designed to endow RAG systems with the ability to respond with "I don't know" when the query is out of the knowledge boundary of both the retrieved passages and the model's internal knowledge. \ourmethod divides data samples into four knowledge quadrants and constructs tailored preference data for each quadrant, resulting in a curated dataset for Direct Preference Optimization (DPO). Experimental results on three benchmark datasets demonstrate that \ourmethod effectively balances accuracy with appropriate abstention, enhancing the reliability and trustworthiness of retrieval-augmented systems.\footnote{Code is available at: \href{https://github.com/JiananXie/Divide-Then-Align}{Divide-Then-Align Repository}}

\end{abstract}

\section{Introduction}

\begin{figure}[t]
  \centering
  \includegraphics[width=0.99\linewidth]{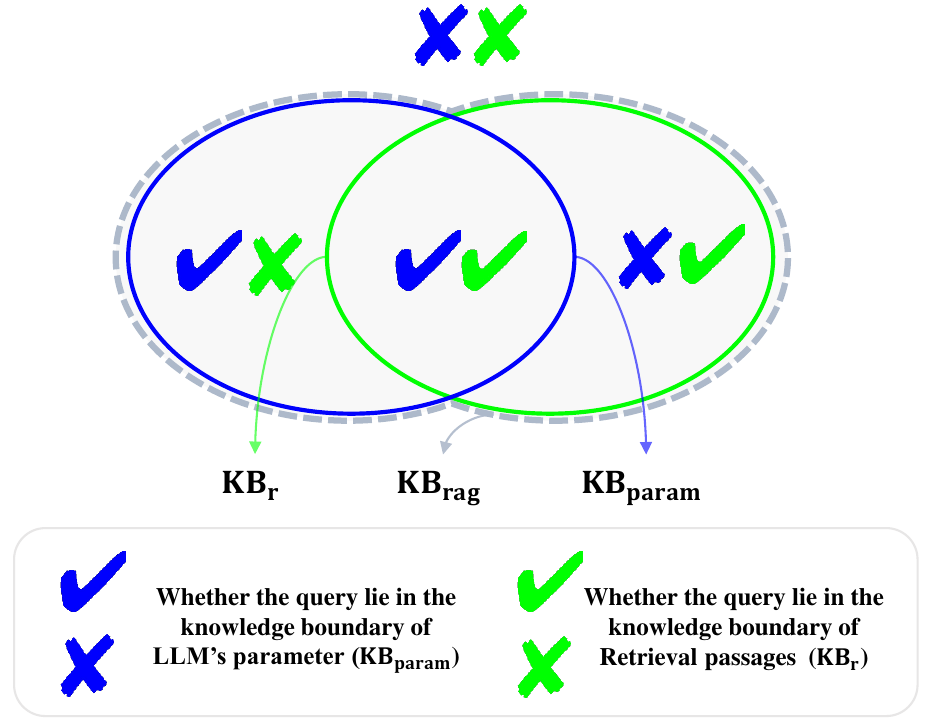}
  \caption{Knowledge Boundary of RAG. A query can be divided into four quadrants based on the model's parametric knowledge boundary ($\KBparam$) and the knowledge boundary of the retrieval passages ($\KBr$). The queries that fall into \bluecross\greencross\ should be answered with "I don't know" instead of generating potentially hallucinatory answers.}
  \label{fig:kb_boundary}
\end{figure}


Large language models (LLMs) have achieved remarkable success across various NLP tasks \cite{radford2019language, brown2020language, bubeck2023sparks, chatgpt}. However, these models are constrained by their pretraining knowledge, which may become outdated or insufficient for domain-specific queries \cite{jiang2023active, shuster-etal-2021-retrieval-augmentation}. Retrieval-Augmented Generation (RAG) \cite{izacard2021leveraging, lewis2020retrieval} addresses this limitation by combining LLMs with retrieval systems that access external knowledge sources \cite{pasca-2019-wikipedia, jin2019pubmedqa} to provide more accurate and contextually rich responses. 

Despite its promise, RAG faces significant challenges due to the limitations of current retrieval systems. In practice, retrieval systems often fail to return entirely accurate passages, resulting in noisy contexts that can contain irrelevant, conflicting, or misleading information \cite{robustlm,fang2024enhancing,cuconasu2024power}. \citet{robustlm, fang2024enhancing,liu2024chatqa} propose Retrieval-Augmented Fine-Tuning (RAFT) to mitigate this issue, which involves fine-tuning LLMs with a combination of retrieved contexts, both relevant and noisy, encouraging the models to learn robustness to noisy inputs.

While RAFT has shown improvements in model performance, it introduces a critical drawback: \textbf{RAFT conditions the model to answer questions even when the retrieved contexts are entirely noisy}. This behavior poses a significant risk for deploying LLMs in real-world applications, particularly in high-stakes domains like medical \cite{raja2024rag}, legal \cite{reji2024enhancing}, and financial \cite{yepes2024financial} fields. As shown in Figure \ref{fig:kb_boundary}, the knowledge boundary of RAG systems is the union of the model's parametric knowledge boundary and the retrieval knowledge boundary. \textbf{When faced with queries for which neither the model's parametric knowledge contains sufficient information to answer the query (\bluecross), nor can useful information be found in the retrieved passages (\greencross), an ideal LLM should respond with "I don't know" instead of generating potentially hallucinatory answers}. However, our experiments reveal that RAFT models do not have this critical ability. Even when explicitly prompted to respond with "I don't know". In such scenarios, the models tend to overfit to the training paradigm and generate hallucinatory answers.

To address this limitation, we propose Divide-Then-Align (\ourmethod), a systematic post-training approach to enhance RAFT models. \ourmethod operates in two key stages: \ding{182} \textbf{Divide}: First, we divide data samples from three benchmark datasets (Natural Questions, TriviaQA, and WebQuestions) into four quadrants based on whether the answers lie within the LLM's parametric knowledge boundary and the retrieval knowledge boundary. This division is crucial as different knowledge quadrants require distinct strategies for preference data construction. \ding{183} \textbf{Align}: For each category, we carefully construct preference data by specifying appropriate chosen and rejected responses based on the knowledge boundary division. This results in a curated training set of 10,000 preference samples. We then employ Direct Preference Optimization (DPO) \cite{rafailov2024direct} to endow the model with the ability to acknowledge uncertainty with "I don't know" responses while maintaining the high accuracy achieved through RAFT training. To rigorously evaluate our approach, we develop \textbf{a comprehensive knowledge quadrants based evaluation framework with nine metrics} that assess both the model's overall performance and its ability to abstain from answering when queries fall outside both knowledge boundaries. Through careful analysis across different quadrants, we demonstrate the effectiveness of our approach in balancing accuracy with principled abstention behavior. 

Our contributions can be summarized as follows:

\begin{itemize}[leftmargin=*]
  \item[\ding{182}] \textbf{Problem Identification}: We first divide the RAG samples into four quadrants based on whether the answers lie within the LLM's parametric knowledge boundary and the retrieval knowledge boundary. And we find that the RAFT model is not able to abstain from answering when the rag sample is out of both the LLM's parametric knowledge boundary and the retrieval knowledge boundary.
  \item[\ding{183}] \textbf{Proposed Solution}: We propose \ourmethod, a systematic approach that constructs quadrant-specific preference data (10,000 samples) and leverages DPO to enable principled abstention behavior while preserving model performance.
  \item[\ding{184}] \textbf{Experimental Validation}: We evaluate our method on three widely used datasets, demonstrating its effectiveness in improving model reliability and trustworthiness.
\end{itemize} 


\section{Preliminary}

\subsection{Knowledge Boundary of RAG}

Let $\mathcal{D}$ denote the knowledge corpus. Let $r: \mathcal{Q} \rightarrow \mathcal{P}$ be the retrieval function that maps a query $q$ to relevant passages $P \subseteq \mathcal{D}$, where $\mathcal{Q}$ is the query space and $\mathcal{P}$ is the passage space. We use $M: \mathcal{Q} \times \mathcal{P} \rightarrow \mathcal{A}$ to represent the LLM function that takes both the query and passages as input and generates an answer from the answer space $\mathcal{A}$. Let $\text{golden}: \mathcal{Q} \rightarrow \mathcal{A}$ be the function that maps a query to its ground truth answer, which represents the correct response that should be generated for the query. Let $C(M(q, P))$ denote the correctness evaluation function.

For honest alignment of RAG systems, it's crucial to determine whether a query q lies within or outside the system's knowledge boundary $\KBrag$. Ideally:

\begin{itemize}
  \item If q $\in \KBrag$, the model should generate the correct answer other than IDK.
  \item If q $\notin \KBrag$, the model should abstain from answering.
\end{itemize}

\subsection{Knowledge Quadrants}

To better evaluate the knowledge boundary of RAG systems, we consider that $\KBrag$ is composed of two fundamental components: the parametric knowledge boundary of the LLM ($\KBparam$) and the knowledge boundary of the retrieval passages ($\KBr$). Formally:


\begin{align}
    \KBparam &= \{q \in \mathcal{Q} \mid C(M(q, \emptyset)) = \text{True}\} \\
    \KBr &= \{q \in \mathcal{Q} \mid \exists p \in r(q): \notag \\
    & \text{contains}(p, \text{golden}(q)) = \text{True}\}
\end{align}

The overall knowledge boundary of the RAG system can be characterized as:
\begin{equation*}
    \KBrag = \KBparam \cup \KBr
\end{equation*}
This formulation captures that a query can be answered correctly if it falls within either the model's parametric knowledge or can be answered using retrieved information.

Then we can divide the samples into quadrants based on \textcolor{blue}{$\KBparam$} and \textcolor{green}{$\KBr$}: 

\begin{itemize}
  \item[\bluecheck\greencheck] : $q \in \KBparam \cap \KBr$ 
  \item[\bluecheck\greencross] : $q \in \KBparam \setminus \KBr$ 
  \item[\bluecross\greencheck] : $q \in \KBr \setminus \KBparam$ 
  \item[\bluecross\greencross] : $q \notin \KBparam \cup \KBr$ 
\end{itemize}

The details of the description of the four quadrants can be found in the Appendix \ref{appendix:knowledge_quadrants}.

\section{Methodology}
\begin{figure*}[t]
    \centering
    \includegraphics[width=1.0\textwidth]{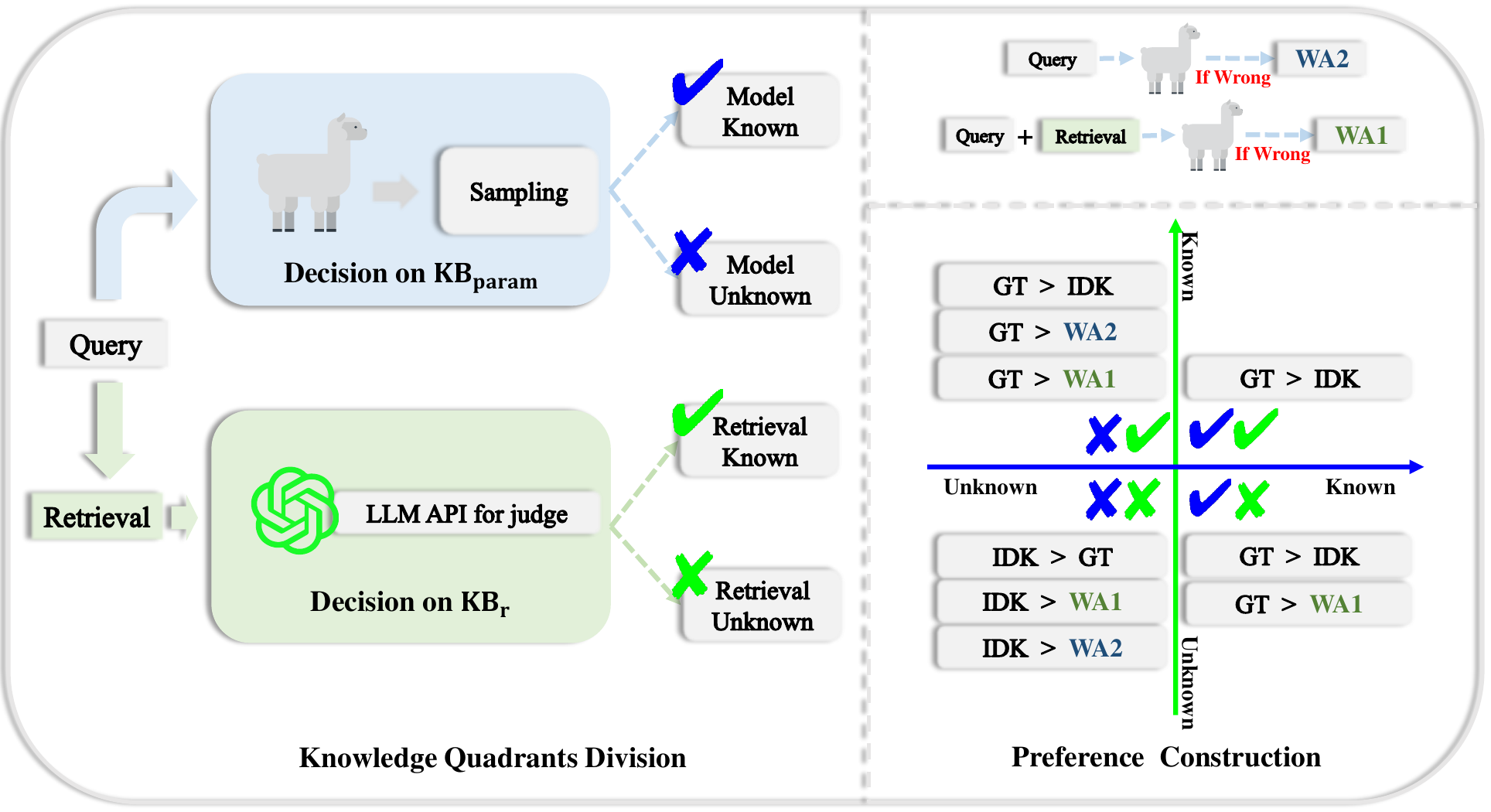}
    \caption{The pipeline of knowledge quadrants division and preference dataset construction. GT denotes the ground truth answer; IDK represents ``I don't know'' response; WA1 and WA2 are wrong answers generated by the LLM (WA = Wrong Answer); ``If Wrong'' indicates the condition where the model generates an incorrect response. The symbol ``>'' indicates a preference relationship where the left option is preferred over the right option. The preference construction (right) shows how different response types (GT, IDK, WA1, WA2) are ranked based on the knowledge quadrant the query falls into. $\KBparam$ means the LLM's parametric knowledge boundary and $\KBr$ means the retrieval knowledge boundary.}
    \label{fig:framework}
\end{figure*}

\subsection{Knowledge Quadrants Division}
To divide queries into the four knowledge quadrants defined in Section 2, we need to determine whether a query $q$ belongs to $\KBparam$ and/or $\KBr$. We use three widely-used question answering datasets: Natural Questions \cite{nq}, TriviaQA \cite{triviaqa}, and WebQuestions \cite{webq}.

\paragraph{Determining $q \in \KBparam$} To determine whe-ther a query lies within the model's parametric knowledge boundary ($q \in \KBparam$), we sample $N$ answers $\{a_1, ..., a_N\}$ from the model without any retrieved context by evaluating $C(M(q, \emptyset))$ with different random seeds. If the proportion of correct answers in these $N$ samples exceeds a threshold
\begin{equation*}   
    \delta = \frac{1}{N}\sum_{i=1}^N \mathbbm{1}[C(a_i)=\text{True}] > \delta
\end{equation*}
we consider $q \in \KBparam$ (\bluecheck). Otherwise, we consider $q \notin \KBparam$ (\bluecross).

To determine whether a response is correct, we directly using lexical matching, which checks whether the golden answers appear in the responses generated by the model. According to the results shown in \cite{wang2024evaluating}, applying lexical matching yields a consisitency rate of approximately 90\% when compared to human evaluation. Therefore, we deem the lexical matching to be a good enough way to determine whether the response is correct. 

\paragraph{Determining $q \in \KBr$} To determine whether a query lies within the retrieval knowledge boundary ($q \in \KBr$), we use GPT-4o (gpt-4o-2024-08-06) to evaluate whether the retrieved passages contain or directly imply the correct answer. We prompt GPT-4o with a specialized evaluation prompt (see Appendix \ref{appendix:prompts}) that returns a binary score indicating whether the context sufficiently supports the answer. If GPT-4o determines the context contains or implies the correct answer (score = 1), we consider $q \in \KBr$ (\greencheck). Otherwise, we consider $q \notin \KBr$ (\greencross).

\subsection{Preference Data Construction}

Based on the knowledge quadrants, we construct preference data for each quadrant as follows:

For \bluecheck\greencheck, we can directly use the ground truth  as the chosen response and use IDK as the rejected response.

For \bluecheck\greencross\ samples, we select the ground truth as the chosen response, while constructing two types of rejected responses: 
(1) incorrect answers generated by the LLM when exposed to noisy context, demonstrating the model's vulnerability to noisy information; and 
(2) "I don't know" responses, which are overly conservative given the model's inherent knowledge.

For \bluecross\greencheck\ samples, the ground truth serves as the chosen response, paired with three categories of rejected responses: 
(1) incorrect answers resulting from the model's failure to utilize the golden information in the context; 
(2) incorrect answers generated by the LLM without any context to suppress the wrong parametric knowledge; and 
(3) "I don't know" responses, which indicate an inability to leverage available context.

For \bluecross\greencross\ samples, where neither source contains reliable information, we designate "I don't know" as the chosen response. The rejected responses comprise: 
(1) incorrect answers generated by the LLM without any context, 
(2) incorrect answers generated by the LLM with noisy context, 
and (3) the ground truth itself, as generating correct answers without supporting evidence may encourage unfounded speculation. 

\paragraph{I don't know Response}  Our refusal to answer template is: 
\begin{tcolorbox}[colback=gray!10,boxrule=0.5pt]
    This question is beyond the scope of my knowledge and the references. I don't know the answer.
\end{tcolorbox}

We use "I don't know" to refer to this template in the paper.

\subsection{Post training using DPO}
In this section, we introduce how to post-train the RAFT model to enable it with the ability to abstain from answering. 

After the preference data is constructed, we employ a multi-objective training approach combining three different losses.

\paragraph{DPO Loss} We utilize the standard DPO loss to learn from preference pairs of chosen and rejected responses. This helps the model learn to distinguish between preferred and non-preferred outputs. Given a chosen response $y_c$ and a rejected response $y_r$ for a query $q$ and retrieved context $r(q)$, the DPO loss is defined as:
\begin{equation}
    \small
    \mathcal{L}_{\text{DPO}} = -\log\sigma(\tau(r_\theta(q,r(q),y_c) - r_\theta(q,r(q),y_r))) 
\end{equation}
where $r_\theta(q,r(q),y)$ represents the log probability of generating response $y$ given query $q$ and retrieved context $r(q)$ under the model parameters $\theta$, $\tau$ is the temperature parameter, and $\sigma$ is the sigmoid function. Note that this reward score is derived from the same language model being trained, eliminating the need for a separate reward model.

\paragraph{SFT Loss} Our empirical observations show that DPO training tends to focus on reducing rejected response rewards rather than improving the quality of the chosen response. To address this limitation, we incorporate supervised fine-tuning loss on the chosen responses to explicitly enhance the model's ability to generate preferred outputs:
\begin{equation}
    \mathcal{L}_{\text{SFT}} = -\sum_{t=1}^T \log p_\theta(y_c^t|q,r(q),y_c^{<t})
\end{equation}
where $y_c^t$ represents the $t$-th token of the chosen response, and $T$ is the length of the response.

\paragraph{Knowledge Quadrant Classification Loss} We add a value head on top of the last token's hidden state to predict which knowledge quadrant (0-3) a query belongs to. This classification task serves as an auxiliary objective that helps the model develop better awareness of its knowledge boundaries and improve its ability to determine when to abstain from answering. The classification loss is defined as:
\begin{equation}
    \mathcal{L}_{\text{class}} = -\sum_{k=0}^3 y_k \log p_{\theta}(k \mid q) \label{eq:loss_class}
\end{equation}
where $y_k$ is the one-hot encoded ground truth label for the knowledge quadrant, and $p_\theta(k|q)$ is the predicted probability for quadrant $k$.

The final training objective is a weighted combination of these three losses:
\begin{equation}
    \mathcal{L}_{\text{total}} = \mathcal{L}_{\text{DPO}} + \beta\mathcal{L}_{\text{SFT}} + \gamma\mathcal{L}_{\text{class}}, \label{eq:loss_total}
\end{equation}
where $\beta$, and $\gamma$ are hyperparameters controlling the contribution of each loss component.

\section{Experiments}
\subsection{Datasets}
We evaluate our approach on three standard open-domain question answering datasets: Natural 
Questions (NQ) \cite{kwiatkowski2019natural}, TriviaQA \cite{joshi2017triviaqa}, and WebQuestions (WebQ) \cite{berant2013semantic}. For each dataset, we follow the setting of \cite{fang2024enhancing} and employ the retrieval model DPR \cite{karpukhin2020dense} as our retriever, which retrieves 3 passages from wikipedia for each query.


To evaluate the model's ability to make appropriate abstentions, we also divide each sample in the test sets into four quadrants based on knowledge boundaries(\bluecheck\greencheck, \bluecheck\greencross, \bluecross\greencheck, \bluecross\greencross). We determine whether a query belongs to the LLM's parametric knowledge ($\KBparam$) based on the performance of vanilla model (LLaMA-2-7b, etal.), and evaluate retrieval knowledge ($\KBr$) based on whether the top-3 retrieved passages contain the correct answer. This division approach allows us to analyze both the RAFT model's improvements over the base model across different knowledge quadrants and its abstention capabilities. After division, we randomly select 3000 queries from three datasets to evaluate all methods.

\label{appendix:dataset}

\begin{table}
  \centering
  \begin{tabular}{l|rrrr}
  \toprule
  \textbf{Dataset} & \multicolumn{1}{c}{\bluecheck\greencheck} & \multicolumn{1}{c}{\bluecheck\greencross} & \multicolumn{1}{c}{\bluecross\greencheck} & \multicolumn{1}{c}{\bluecross\greencross} \\
  \midrule
  \multicolumn{5}{l}{\textit{LLaMA-2-7B}} \\
  \midrule
  NQ & 204 & 40 & 2,125 & 1,241 \\
  TriviaQA & 2,225 & 1,109 & 4,391 & 3,588 \\
  WebQ & 202 & 76 & 882 & 872 \\
  \midrule
  \multicolumn{5}{l}{\textit{LLaMA-2-13B}} \\
  \midrule
  NQ & 451 & 105 & 1,877 & 1,172 \\
  TriviaQA & 3,669 & 1,978 & 2,809 & 2,652 \\
  WebQ & 258 & 105 & 826 & 843 \\
  \midrule
  \multicolumn{5}{l}{\textit{LLaMA-3-8B}} \\
  \midrule
  NQ & 442 & 122 & 1,887 & 1,159 \\
  TriviaQA & 3,229 & 1,721 & 3,387 & 2,976 \\
  WebQ & 224 & 94 & 860 & 854 \\
  \bottomrule
  \end{tabular}
  \caption{Statistics of the test set across different model architectures and datasets. The columns show the distribution of samples across the four knowledge quadrants.}
  \label{tab:testset}
  \end{table}

To balance the model's ability to answer questions and abstain when appropriate, we introduce a hyperparameter called IDK-ratio, which controls the proportion of training examples where the preferred response is "I don't know" (IDK). Specifically, IDK-ratio determines the fraction of \bluecross\greencross\ samples in the training set. Importantly, we maintain the natural distribution of queries across all four quadrants in the test set without any manipulation, ensuring evaluation reflects real-world conditions and provides a more generalizable assessment of model performance.


Table \ref{tab:testset} shows the distribution of test queries across the four knowledge quadrants. A substantial portion of queries fall into the
 \bluecross\greencross\ quadrant. This represents a critical scenario where models should abstain from answering, yet traditional RAFT approaches force a response. The distribution highlights why defining $\KBrag$ through the combination of both $\KBparam$ and $\KBr$ is crucial. Relying solely on $\KBr$ \cite{liu2024chatqa,song2024measuring} would incorrectly exclude \bluecheck\greencross\ queries from the model's knowledge boundary (for example, 1,978 TriviaQA queries for LLaMA-2-13B where the model has parametric knowledge). Similarly, using only $\KBparam$ \cite{cheng2024can,feng-etal-2024-dont,xu2024rejection} would mistakenly omit \bluecross\greencheck\ queries (such as 2,125 NQ queries for LLaMA-2-7B) that RAG systems can effectively handle through retrieval. Our dual-boundary approach enables more precise identification of true knowledge gaps (\bluecross\greencross\ cases) where abstention is warranted, while allowing optimal knowledge source selection in other cases.

 \begin{table*}[h]
  \centering
  \small
  \begin{tabular}{c|c|c|>{\scriptsize}c}
  \toprule
  \textbf{Category} & \textbf{Metric} & \multicolumn{1}{c|}{\textbf{Formula}} & \textbf{Description} \\
  \midrule
  \makecell[c]{Overall\\Quality}
  &  Accuracy & $\frac{|\faCheckMath \cap (\bluecheckMath\greencheckMath \cup \bluecheckMath\greencrossMath \cup \bluecrossMath\greencheckMath)|  +  |\faCircleOMath \cap \bluecrossMath\greencrossMath|}{|\bluecheckMath\greencheckMath \cup \bluecheckMath\greencrossMath \cup \bluecrossMath\greencheckMath \cup \bluecrossMath\greencrossMath|}$ & 
  Ratio of correct answers plus proper abstentions to total queries \\
  \midrule
  \multirow{3}{*}[-2ex]{\makecell{Answer\\Quality}} 
  & Recall & $\frac{|\faCheckMath \cap (\bluecheckMath\greencheckMath \cup \bluecheckMath\greencrossMath \cup \bluecrossMath\greencheckMath)|}{|\bluecheckMath\greencheckMath \cup \bluecheckMath\greencrossMath \cup \bluecrossMath\greencheckMath|}$ & 
  Ratio of correct answers to all queries in $\KBrag$ \\
  \cmidrule{2-4}
  & Precision & $\frac{|\faCheckMath \cap (\bluecheckMath\greencheckMath \cup \bluecheckMath\greencrossMath \cup \bluecrossMath\greencheckMath)|}{|\faCheckMath| + |\faCloseMath|}$ & 
  Ratio of correct answers to attempted answers \\
  \cmidrule{2-4}
  & F1 & $\frac{2 \cdot \text{Prec} \cdot \text{Rec}}{\text{Prec} + \text{Rec}}$ & 
  The harmonic mean of precision and recall \\
  \midrule
  \multirow{2}{*}[-1ex]{\makecell[c]{Retrieval\\Handling}} 
    & \footnotesize{Denoise Rate} & $\frac{|\faCheckMath \cap \bluecheckMath\greencrossMath|}{|\bluecheckMath\greencrossMath|}$ & 
  Ability to ignore noisy retrieval \\
  \cmidrule{2-4}
  & \footnotesize{Context Utilization Rate} & $\frac{|\faCheckMath \cap \bluecrossMath\greencheckMath|}{|\bluecrossMath\greencheckMath|}$ & 
  Ability to utilize golden information \\
  \midrule
  \multirow{3}{*}[-2ex]{\makecell[c]{Abstain\\Quality}}
  & \footnotesize{Abstain Recall} & $\frac{|\faCircleOMath \cap \bluecrossMath\greencrossMath|}{|\bluecrossMath\greencrossMath|}$ & 
  Ratio of correct abstentions to all queries in $\bluecrossMath\greencrossMath$ \\
  \cmidrule{2-4}
  & \footnotesize{Abstain Precision} & $\frac{|\faCircleOMath \cap \bluecrossMath\greencrossMath|}{|\faCircleOMath|}$ & 
  Ratio of correct abstentions to all abstentions \\
  \cmidrule{2-4}
  & \footnotesize{Abstain F1} &  $\frac{2 \cdot \text{AbPrec} \cdot \text{AbRec}}{\text{AbPrec} + \text{AbRec}}$ & 
  The harmonic mean of abstain precision and abstain recall \\
  \bottomrule
  \end{tabular}
  \caption{Evaluation Metrics based on the knowledge quadrant division. Let $\faCheckMath$ denote correct answers, $\faCloseMath$ denote incorrect answers, and $\faCircleOMath$ denote abstentions ("I don't know" responses). For any category (e.g., 
  $\bluecheckMath\greencrossMath$), $|\faCheckMath \cap \bluecheckMath\greencrossMath|$ represents the count of correct answers within the 
  \bluecheckMath\greencrossMath\ category.}
  \label{tab:metrics}
  \end{table*}

\subsection{Baselines}

We evaluate our approach against three categories of baselines: (1) RAFT models that focus on handling retrieval noise (\textbf{RAAT} \cite{fang2024enhancing}, \textbf{Ret-Robust} \cite{robustlm}, \textbf{ChatQA} \cite{liu2024chatqa}),  (2) calibration-based methods that detect potential hallucinations (\textbf{P(True)} \cite{kadavath2022language}, \textbf{Logits} \cite{guerreiro-etal-2023-looking}) and (3) two widely-used baselines like in-context learning (\textbf{ICL} \cite{wei2022chain}) and  \textbf{self-Consistency} \cite{wang2022self}.  Details of these baselines can be found in Appendix \ref{appendix:baselines} and \ref{appendix:baselines_implementation}.

\subsection{Evaluation Metrics}
To systematically evaluate the performance of our method, we propose a comprehensive evaluation framework based on the knowledge quadrant division. The framework consists of four main aspects: Overall Quality (OQ), Answer Quality (AQ), Retrieval Handling (RH), and Abstention Quality (AbQ). Across these aspects, we define 9 distinct metrics that thoroughly assess different dimensions of model performance. The details and formulations of these metrics are presented in Table \ref{tab:metrics}.

\begin{table*}[t]
  \centering
  \begin{tabular}{ll|ccccccccc}
  \toprule
  & & OQ & \multicolumn{3}{c}{AQ} & \multicolumn{2}{c}{RH} & \multicolumn{3}{c}{AbQ} \\
  \cmidrule(lr){3-3} \cmidrule(lr){4-6} \cmidrule(lr){7-8} \cmidrule(lr){9-11}
  & Model Name & Acc & Rec & Prec & F1 & DR & CUR & ARec & APrec & AF1 \\ \midrule
  \multicolumn{11}{c}{\textit{Llama-2-7b}} \\\midrule
  & Original    & 42.2 & 64.1 & 42.2 & 50.9 & \textbf{85.8} & 49.9 & 0.00 & 0.00 & 0.00 \\
  & RAAT        & 46.2 & \textbf{70.2} & 46.2 & 55.7 & 76.3 & \textbf{61.7} & 0.00 & 0.00 & 0.00 \\
  & \quad$+$ P(true)     & 45.0 & 65.0 & 46.0 & 53.8 & 68.9 & 57.4 & 6.71 & 32.1 & 11.0 \\
  & \quad$+$ Logits      & 49.2 & 58.8 & 50.5 & 54.3 & 69.8 & 47.0 & 30.9 & 45.1 & 36.6 \\
  & \quad$+$ Consistency & 51.4 & 69.0 & 50.7 & 58.5 & 82.1 & 58.8 & 16.3 & 58.4 & 25.4 \\
  & \quad$+$ ICL         & 46.8 & 71.2 & 46.8 & 56.5 & 84.4 & 60.2 & 0.00 & 0.00 & 0.00 \\
  \rowcolor{gray!15}
  & \quad$+$ \ourmethod     & \textbf{64.1} & 63.7 & \textbf{65.5} & \textbf{64.6} & 68.9 & 52.8 & \textbf{65.0} & \textbf{61.7} & \textbf{63.3} \\ \midrule
  \multicolumn{11}{c}{\textit{Llama-2-13b}} \\ \midrule
  & Original    & 48.1 & 66.3 & 48.1 & 55.8 & 82.1 & 40.7 & 0.00 & 0.00 & 0.00 \\
  & Ret-Robust  & 51.6 & 71.0 & 51.6 & 59.8 & \textbf{90.0} & 44.5 & 0.00 & 0.00 & 0.00 \\
  & \quad$+$ P(true)     & 50.9 & 56.0 & 58.5 & 57.2 & 74.8 & 29.7 & 37.5 & 33.6 & 35.4 \\
  & \quad$+$ Logits      & 53.6 & 70.0 & 53.6 & 60.7 & 87.9 & 43.4 & 10.0 & 52.9 & 16.9 \\
  & \quad$+$ Consistency & 53.9 & \textbf{71.8} & 54.0 & 61.7 & 89.6 & \textbf{46.4} & 6.30 & 52.5 & 11.2 \\
  & \quad$+$ ICL         & 52.0 & 71.6 & 52.0 & 60.3 & 89.1 & 46.6 & 0.00 & 0.00 & 0.00 \\
  \rowcolor{gray!15}
  & \quad$+$ \ourmethod        & \textbf{64.8} & 67.9 & \textbf{65.3} & \textbf{66.6} & 76.8 & 45.5 & \textbf{56.7} & \textbf{63.5} & \textbf{59.9} \\ \midrule
  \multicolumn{11}{c}{\textit{Llama-3-8b}} \\ \midrule
  & Original    & 43.9 & 62.0 & 43.9 & 51.4 & \textbf{76.0} & 42.0 & 0.00 & 0.00 & 0.00 \\
  & ChatQA      & 46.1 & 60.9 & 45.0 & 51.8 & 54.5 & 46.8 & 10.2 & 71.8 & 17.8 \\
  & \quad$+$ P(true)     & 50.1 & 45.2 & 55.6 & 49.9 & 46.2 & 29.1 & 61.9 & 42.6 & 50.5 \\
  & \quad$+$ Logits      & 46.6 & 57.8 & 46.8 & 51.7 & 51.0 & 44.8 & 19.3 & 44.9 & 27.0 \\
  & \quad$+$ Consistency & 46.5 & 61.0 & 46.7 & 52.9 & 58.7 & 46.6 & 11.3 & 44.0 & 18.0 \\
  & \quad$+$ ICL         & 43.3 & 55.0 & 41.4 & 47.2 & 50.3 & 40.7 & 15.1 & \textbf{75.4} & 25.1 \\
  \rowcolor{gray!15}
  & \quad$+$ \ourmethod        & \textbf{65.5} & \textbf{64.5} & \textbf{67.2} & \textbf{65.8} & 62.8 & \textbf{48.9} & \textbf{67.9} & 61.8 & \textbf{64.7} \\ \bottomrule
  \end{tabular}
  \caption{Main results on the benchmark consisting of three datasets. OQ: Overall Quality (Acc: Accuracy); AQ: Answer Quality (Rec: Recall, Prec: Precision); RH: Retrieval Handling (DR: Denoise Rate, CUR: Context Utilization Rate); AbQ: Abstain Quality (ARec: Abstain Recall, APrec: Abstain Precision, AF1: Abstain F1).}
\label{tab:main}
  \end{table*}


\begin{table*}[h]
  \centering
  \begin{tabular}{c|ccccccccc}
  \toprule
  & OQ & \multicolumn{3}{c}{AQ} & \multicolumn{2}{c}{RH} & \multicolumn{3}{c}{AbQ} \\
  \cmidrule(lr){2-2} \cmidrule(lr){3-5} \cmidrule(lr){6-7} \cmidrule(lr){8-10}
  Model Name & Acc & Rec & Prec & F1 & DR & CUR & ARec & APrec & AF1 \\ 
  \midrule
  \ourmethod  & 64.1 & 63.7 & 65.5 & 64.6 & 68.9 & 52.8 & 65.0 & 61.7 & 63.3\\
  \midrule
  w/o DPO & 52.4 & 38.8 & 67.8 & 49.4 & 52.1 & 28.7 & 78.6 & 43.1 & 55.7 \\
  w/o SFT & 37.1 & 54.6 & 36.5 & 43.8 & 58.9 & 45.2 & 3.50 & 76.6 & 6.7 \\
  w/o CLS & 63.1 & 63.5 & 63.3 & 63.4 & 63.9 & 53.6 & 62.4 & 62.7 & 62.6 \\
  \midrule
  w/o \bluecheck\greencheck & 57.0 & 54.6 & 59.0 & 56.7 & 57.1 & 43.9 & 61.5 & 53.9 & 57.5 \\
  w/o \bluecheck\greencross & 61.7 & 53.4 & 67.3 & 59.5 & 47.9 & 44.5 & 77.7 & 55.7 & 64.9 \\
  w/o \bluecross\greencheck & 58.6 & 58.5 & 59.8 & 59.1 & 72.1 & 45.6 & 58.7 & 56.5 & 57.6 \\
  w/o \bluecross\greencross & 48.2 & 73.3 & 48.2 & 58.2 & 84.5 & 64.0 & 0.00 & 0.00 & 0.00 \\
  \midrule
  w/o WA1 & 61.8 & 68.8 & 59.0 & 63.5 & 75.3 & 58.8 & 48.4 & 71.2 & 57.6 \\
  w/o WA2 & 61.5 & 66.2 & 59.1 & 62.4 & 68.5 & 56.4 & 52.4 & 68.2 & 59.3 \\
  w/o WA1$\cup$WA2 & 58.2 & 68.5 & 53.8 & 60.3 & 71.7 & 59.4 & 38.5 & 80.6 & 52.1 \\
  \bottomrule
  \end{tabular}
  \caption{Ablation results.}
\label{tab:ablation}
\end{table*}

\subsection{Main Results}
Main experimental results are shown in Table \ref{tab:main}. Our post-training strategy \ourmethod achieves the best performance on three llama architectures. Notably, it achieves Acc (64.1, 64.8, 65.5), F1 (64.6, 66.6 65.8), AF1(63.3, 59.9, 64.7), surpassing baseline methods by significant margins. Critically, \ourmethod uniquely balances robust answer generation with precise abstention, addressing a key limitation of existing approaches.

While RAFT variants (RAAT, Ret-Robust, ChatQA) can improve answer quality of base model, they uniformly fail to abstain properly. As designed, RAFT models effectively enhance the model answer quality. In addition, following its training approach, RAAT did a good job of using golden contexts to generate correct answers. Ret-Robust can resist the most noisy retrieval and generate high-quality responses using model's knowledge. However, they all struggle with abstain quality. In both RAAT and Ret-Robust, none of the test queries can be abstained. ChatQA has the ability to refrain from some queries, but the quality is far from satisfactory. Post-hoc techniques, including two calibration methods (P(true), Logits) and consistency, are applied to RAFT models to enhance abstain quality but impair the ability to use model knowledge. And their answer quality is also affected, which is not good for the overall performance. ICL only improves the abstain quality when the RAFT model has the ability to abstain, but the improvement is not significant. 

In stark contrast, \ourmethod achieves highest AF1 without compromising answer quality. DTA did this by structurally aligning model behavior with knowledge boundaries, enabling reliable and self-aware QA systems. However, our method falls short in terms of the DR and CUR metrics, which is related to the trade-off with abstention. When appropriately enhancing the model's abstention capability to promote the growth of overall quality, a significant portion of the \bluecheck\greencross\ and \bluecross\greencheck\ data is also rejected. On the contrary, a significant reduction in the proportion of \bluecross\greencross\ during training leads to a notable surge in both DR and CUR scores. Further discussion is shown in hyperparamter experiments. 

An interesting observation is that the original LLM achieves remarkably high DR scores. While RAFT models are specifically trained to utilize context and rely more heavily on retrieved passages for generating answers, recent research \cite{tan2024blinded, bi2024context} suggests that base models tend to prioritize their parametric knowledge while being less dependent on provided context. Since all contexts in the DR category are noisy, excessive reliance on context would only lead to degraded performance. 

To better understand the impact of knowledge quadrant division, we conducted experiments using single knowledge boundaries ($\KBr$ or $\KBparam$) instead of the full quadrant approach. For these experiments, we used ground truth answers when queries fell within the knowledge boundary and abstention responses when queries fell outside it, while keeping all other hyperparameters identical to \ourmethod. As shown in Table \ref{tab:KB}, using single knowledge boundaries led to notably worse performance across metrics, demonstrating the importance of our fine-grained quadrant-based approach for properly modeling RAG system knowledge boundaries.

\begin{table*}[h]
  \centering
  \begin{tabular}{c|ccccccccc}
  \toprule
    & OQ &\multicolumn{3}{c}{AQ} & \multicolumn{2}{c}{RH} & \multicolumn{3}{c}{AbQ} \\
  \cmidrule(lr){2-2} \cmidrule(lr){3-5} \cmidrule(lr){6-7} \cmidrule(lr){8-10}
   Knowledge Boundary & Acc & Rec & Prec & F1 & DR & CUR & ARec & APrec & AF1 \\ 
  \midrule
   \ourmethod  & 64.1 & 63.7 & 65.5 & 64.6 & 68.9 & 52.8 & 65.0 & 61.7 & 63.3\\
   $\KBr$ & 58.9 & 49.4 & 62.9 & 55.3 & 43.4 & 41.7 & 77.3 & 54.7 & 64.1 \\
 $\KBparam$ & 45.8 & 32.6 & 42.6 & 36.9 & 39.3 & 23.1 & 71.1 & 49.0 & 58.0 \\
  \bottomrule
  \end{tabular}
  \caption{Experimental results on different knowledge boundary.}
\label{tab:KB}
  \end{table*}

\subsection{Ablation Study}
We conducted comprehensive ablation experiments to analyze the contribution of each component in our \ourmethod framework based on the \ourmethod results of RAAT. The results in Table \ref{tab:ablation} demonstrate the importance of each component from multiple aspects:

\paragraph{Training Objectives} Without DPO loss, the model shows significantly degraded performance in answer quality (Rec drops from 63.7\% to 38.8\%) while maintaining high abstention rates (ARec: 78.6\%). However, the abstain precision decreases substantially from 61.7\% to 43.1\%. This indicates that although the RAG system learns to abstain, it becomes overly cautious and lacks confidence in answering queries that it should be able to handle. Without SFT loss, the model exhibits a dramatic decline in overall quality (Acc drops from 63.7\% to 38.8\%) and severely degraded abstention quality (AF1 drops from 63.3\% to 6.7\%). These results validate our hypothesis that the SFT loss plays a crucial role in teaching the model how to make abstention. The removal of classification loss shows relatively minor impact across metrics, with slight decreases in both answer quality (F1 drops from 64.6\% to 63.4\%) and abstention quality (AF1 drops from 63.3\% to 62.6\%). This suggests that while knowledge quadrant classification serves as a helpful auxiliary task, it is not critical to the model's core capabilities.

\paragraph{Knowledge Boundary Components} Removing \bluecheck\greencheck\ samples from training leads to decreased performance across all metrics, particularly in context utilization (CUR drops to 43.9\%), highlighting the importance of learning from samples where correct information is available in the context. Without \bluecheck\greencross\ samples, the model shows reduced ability to handle retrieved information (DR: 47.9\%), indicating that exposure to noisy samples during training is crucial for developing robust retrieval handling capabilities. Without \bluecross\greencheck\ samples, the model shows an interesting trade-off: while the denoise rate (DR) improves to 72.1\%, the context utilization rate (CUR) drops to 45.6\%. This suggests that without training on samples where the model needs to rely on retrieved context, it becomes overly conservative with retrieval usage, preferring to rely on its parametric knowledge even when helpful context is available. This leads to degraded overall accuracy (58.6\%), highlighting the importance of these samples for teaching the model when to effectively leverage retrieved information. Without \bluecross\greencross\ samples, the model completely loses its abstention capability (AbQ metrics all 0.0) while showing artificially high recall (73.3\%) and DR (84.5\%), indicating that training with examples where abstention is appropriate is essential for developing proper abstention behavior.

\paragraph{Wrong Answer Types} The impact of removing wrong answer types (w/o WA1, w/o WA2) reveals an interesting trade-off in model behavior. Without the suppression of wrong answers, the model becomes more inclined to generate responses rather than abstain, leading to higher recall (68.8\% for w/o WA1, 66.2\% for w/o WA2) and improved retrieval handling metrics. However, this increased response rate comes at the cost of precision, dropping from 65.5\% to around 59\%, as the total number of attempted answers grows significantly. The model's abstention capability is also compromised, with lower abstention recall but higher abstention precision, indicating more conservative use of "I don't know" responses. These results demonstrate that wrong answer samples play a crucial role in training by helping the model establish appropriate decision boundaries between answering and abstaining, ultimately contributing to better overall performance when both types are included.


\subsection{Hyperparameter}
Experiments are conducted on  preference dataset size, multi-objective loss weights and IDK-ratio for the preference dataset. The experimental results are shown in Appendix \ref{appendix:hyperparameter}.

\section{Conclusion}

In this paper, we propose a novel framework for honest alignment of retrieval-augmented language models based on knowledge boundary quadrants. We first identify that the knowledge boundary of RAG systems consists of two fundamental components: the parametric knowledge boundary ($\KBparam$) and the retrieval knowledge boundary ($\KBr$). Based on this insight, we divide RAG samples into four knowledge quadrants.  To address the critical limitation of RAFT models regarding their inability to abstain from answering when queries fall outside both knowledge boundaries (\bluecross\greencross), we construct a comprehensive preference dataset that captures the desired behavior for each quadrant. Using this dataset, we employ DPO training with a multi-objective approach combining DPO loss, SFT loss, and knowledge quadrant classification loss to align the model's behavior with the knowledge boundary constraints. Furthermore, we introduce a systematic evaluation framework with 9 metrics to assess both response quality and abstention capabilities. Experiments conducted on three benchmark datasets demonstrate that our approach effectively improves the model's ability to make appropriate abstention decisions while maintaining strong performance on answerable queries.


\section*{Limitations}

While our work presents a promising approach for honest alignment of RAG systems, following limitations should be noted:

\paragraph{Knowledge Boundary Determination}: Our method for determining whether a query belongs to $\KBparam$ relies on sampling from the base model without context, which is used by a lot of previous works \cite{xu2024rejection, cheng2024can}. However, this approach may not perfectly capture the true parametric knowledge boundary, as model performance can vary across different prompting strategies. And we think this is a potential research direction for future work.

\paragraph{Specific Domain}: Our evaluation focuses on three general-domain open QA datasets (NQ, TriviaQA, WebQ). While these datasets provide a good foundation for testing, they may not fully represent the challenges and nuances specific to specialized domain applications. The effectiveness of our approach in highly specialized domains requires further investigation.

\section*{Ethical Considerations}
Our work improves the refusal capability of RAG systems to reduce the risk of generating harmful or incorrect information. Nevertheless, the model may still produce low-quality or hallucinated responses, when faced with ambiguous or out-of-distribution queries. Additionally, since our model has not undergone safety alignment, it may still generate inappropriate content when faced with adversarial or malicious queries.


\section*{Acknowledgments}
This work is sponsored by National Natural Science Foundation of China (62236010, 62141608, 62206291). 


\bibliography{custom, anthology}

\begin{thebibliography}{68}
\providecommand{\natexlab}[1]{#1}

\bibitem[{Asai et~al.(2023)Asai, Wu, Wang, Sil, and Hajishirzi}]{asai2023self}
Akari Asai, Zeqiu Wu, Yizhong Wang, Avirup Sil, and Hannaneh Hajishirzi. 2023.
\newblock Self-rag: Learning to retrieve, generate, and critique through self-reflection.
\newblock \emph{arXiv preprint arXiv:2310.11511}.

\bibitem[{Asai et~al.(2024)Asai, Wu, Wang, Sil, and Hajishirzi}]{asai2024selfrag}
Akari Asai, Zeqiu Wu, Yizhong Wang, Avirup Sil, and Hannaneh Hajishirzi. 2024.
\newblock \href {https://openreview.net/forum?id=hSyW5go0v8} {Self-{RAG}: Learning to retrieve, generate, and critique through self-reflection}.
\newblock In \emph{ICLR}.

\bibitem[{Azaria and Mitchell(2023)}]{azaria2023internal}
Amos Azaria and Tom Mitchell. 2023.
\newblock The internal state of an llm knows when it’s lying.
\newblock In \emph{Findings of the Association for Computational Linguistics: EMNLP 2023}, pages 967--976.

\bibitem[{Berant et~al.(2013{\natexlab{a}})Berant, Chou, Frostig, and Liang}]{webq}
Jonathan Berant, Andrew Chou, Roy Frostig, and Percy Liang. 2013{\natexlab{a}}.
\newblock \href {https://aclanthology.org/D13-1160} {Semantic parsing on {F}reebase from question-answer pairs}.
\newblock In \emph{Proceedings of the 2013 Conference on Empirical Methods in Natural Language Processing}, pages 1533--1544, Seattle, Washington, USA. Association for Computational Linguistics.

\bibitem[{Berant et~al.(2013{\natexlab{b}})Berant, Chou, Frostig, and Liang}]{berant2013semantic}
Jonathan Berant, Andrew Chou, Roy Frostig, and Percy Liang. 2013{\natexlab{b}}.
\newblock \href {https://aclanthology.org/D13-1160} {Semantic parsing on {F}reebase from question-answer pairs}.
\newblock In \emph{Proceedings of the 2013 Conference on Empirical Methods in Natural Language Processing}, pages 1533--1544, Seattle, Washington, USA. Association for Computational Linguistics.

\bibitem[{Bi et~al.(2024{\natexlab{a}})Bi, Huang, Wang, Yang, Zhang, Huang, Mei, Fang, Li, Wei et~al.}]{bi2024context}
Baolong Bi, Shaohan Huang, Yiwei Wang, Tianchi Yang, Zihan Zhang, Haizhen Huang, Lingrui Mei, Junfeng Fang, Zehao Li, Furu Wei, et~al. 2024{\natexlab{a}}.
\newblock Context-dpo: Aligning language models for context-faithfulness.
\newblock \emph{arXiv preprint arXiv:2412.15280}.

\bibitem[{Bi et~al.(2024{\natexlab{b}})Bi, Liu, Wang, Mei, Fang, Gao, Ni, and Cheng}]{bi2024factuality}
Baolong Bi, Shenghua Liu, Yiwei Wang, Lingrui Mei, Junfeng Fang, Hongcheng Gao, Shiyu Ni, and Xueqi Cheng. 2024{\natexlab{b}}.
\newblock Is factuality enhancement a free lunch for llms? better factuality can lead to worse context-faithfulness.
\newblock \emph{arXiv preprint arXiv:2404.00216}.

\bibitem[{Bi et~al.(2025)Bi, Liu, Wang, Xu, Fang, Mei, and Cheng}]{bi2025parameters}
Baolong Bi, Shenghua Liu, Yiwei Wang, Yilong Xu, Junfeng Fang, Lingrui Mei, and Xueqi Cheng. 2025.
\newblock Parameters vs. context: Fine-grained control of knowledge reliance in language models.
\newblock \emph{arXiv preprint arXiv:2503.15888}.

\bibitem[{Borgeaud et~al.(2022)Borgeaud, Mensch, Hoffmann, Cai, Rutherford, Millican, van~den Driessche, Lespiau, Damoc, Clark, de~Las~Casas, Guy, Menick, Ring, Hennigan, Huang, Maggiore, Jones, Cassirer, Brock, Paganini, Irving, Vinyals, Osindero, Simonyan, Rae, Elsen, and Sifre}]{borgeaud2022improving}
Sebastian Borgeaud, Arthur Mensch, Jordan Hoffmann, Trevor Cai, Eliza Rutherford, Katie Millican, George van~den Driessche, Jean{-}Baptiste Lespiau, Bogdan Damoc, Aidan Clark, Diego de~Las~Casas, Aurelia Guy, Jacob Menick, Roman Ring, Tom Hennigan, Saffron Huang, Loren Maggiore, Chris Jones, Albin Cassirer, Andy Brock, Michela Paganini, Geoffrey Irving, Oriol Vinyals, Simon Osindero, Karen Simonyan, Jack~W. Rae, Erich Elsen, and Laurent Sifre. 2022.
\newblock \href {https://proceedings.mlr.press/v162/borgeaud22a.html} {Improving language models by retrieving from trillions of tokens}.
\newblock In \emph{International Conference on Machine Learning, {ICML} 2022, 17-23 July 2022, Baltimore, Maryland, {USA}}, volume 162 of \emph{Proceedings of Machine Learning Research}, pages 2206--2240. {PMLR}.

\bibitem[{Brown et~al.(2020)Brown, Mann, Ryder, Subbiah, Kaplan, Dhariwal, Neelakantan, Shyam, Sastry, Askell et~al.}]{brown2020language}
Tom Brown, Benjamin Mann, Nick Ryder, Melanie Subbiah, Jared~D Kaplan, Prafulla Dhariwal, Arvind Neelakantan, Pranav Shyam, Girish Sastry, Amanda Askell, et~al. 2020.
\newblock Language models are few-shot learners.
\newblock \emph{Advances in neural information processing systems}, 33:1877--1901.

\bibitem[{Bubeck et~al.(2023)Bubeck, Chandrasekaran, Eldan, Gehrke, Horvitz, Kamar, Lee, Lee, Li, Lundberg et~al.}]{bubeck2023sparks}
S{\'e}bastien Bubeck, Varun Chandrasekaran, Ronen Eldan, Johannes Gehrke, Eric Horvitz, Ece Kamar, Peter Lee, Yin~Tat Lee, Yuanzhi Li, Scott Lundberg, et~al. 2023.
\newblock Sparks of artificial general intelligence: Early experiments with gpt-4.
\newblock \emph{arXiv preprint arXiv:2303.12712}.

\bibitem[{Cheng et~al.(2024)Cheng, Sun, Liu, Zhang, Yin, Li, Li, Chen, and Qiu}]{cheng2024can}
Qinyuan Cheng, Tianxiang Sun, Xiangyang Liu, Wenwei Zhang, Zhangyue Yin, Shimin Li, Linyang Li, Kai Chen, and Xipeng Qiu. 2024.
\newblock Can ai assistants know what they don't know?
\newblock In \emph{Forty-first International Conference on Machine Learning}.

\bibitem[{Cuconasu et~al.(2024)Cuconasu, Trappolini, Siciliano, Filice, Campagnano, Maarek, Tonellotto, and Silvestri}]{cuconasu2024power}
Florin Cuconasu, Giovanni Trappolini, Federico Siciliano, Simone Filice, Cesare Campagnano, Yoelle Maarek, Nicola Tonellotto, and Fabrizio Silvestri. 2024.
\newblock The power of noise: Redefining retrieval for rag systems.
\newblock In \emph{Proceedings of the 47th International ACM SIGIR Conference on Research and Development in Information Retrieval}, pages 719--729.

\bibitem[{Duan et~al.(2024)Duan, Cheng, Wang, Zavalny, Wang, Xu, Kailkhura, and Xu}]{duan2024shifting}
Jinhao Duan, Hao Cheng, Shiqi Wang, Alex Zavalny, Chenan Wang, Renjing Xu, Bhavya Kailkhura, and Kaidi Xu. 2024.
\newblock Shifting attention to relevance: Towards the predictive uncertainty quantification of free-form large language models.
\newblock In \emph{Proceedings of the 62nd Annual Meeting of the Association for Computational Linguistics (Volume 1: Long Papers)}, pages 5050--5063.

\bibitem[{Fang et~al.(2024)Fang, Bai, Ni, Yang, Chen, and Xu}]{fang2024enhancing}
Feiteng Fang, Yuelin Bai, Shiwen Ni, Min Yang, Xiaojun Chen, and Ruifeng Xu. 2024.
\newblock Enhancing noise robustness of retrieval-augmented language models with adaptive adversarial training.
\newblock \emph{arXiv preprint arXiv:2405.20978}.

\bibitem[{Feng et~al.(2024)Feng, Shi, Wang, Ding, Balachandran, and Tsvetkov}]{feng-etal-2024-dont}
Shangbin Feng, Weijia Shi, Yike Wang, Wenxuan Ding, Vidhisha Balachandran, and Yulia Tsvetkov. 2024.
\newblock \href {https://doi.org/10.18653/v1/2024.acl-long.786} {Don{'}t hallucinate, abstain: Identifying {LLM} knowledge gaps via multi-{LLM} collaboration}.
\newblock In \emph{Proceedings of the 62nd Annual Meeting of the Association for Computational Linguistics (Volume 1: Long Papers)}, pages 14664--14690, Bangkok, Thailand. Association for Computational Linguistics.

\bibitem[{Gao et~al.(2024)Gao, Zhang, Chen, Huang, Wu, Fu, Wan, Zhang, and Sun}]{gao2024best}
Chujie Gao, Qihui Zhang, Dongping Chen, Yue Huang, Siyuan Wu, Zhengyan Fu, Yao Wan, Xiangliang Zhang, and Lichao Sun. 2024.
\newblock The best of both worlds: Toward an honest and helpful large language model.
\newblock \emph{arXiv preprint arXiv:2406.00380}.

\bibitem[{Ge et~al.(2025)Ge, Liu, Wang, Mei, Chen, Bi, and Cheng}]{ge2025innate}
Yuyao Ge, Shenghua Liu, Yiwei Wang, Lingrui Mei, Lizhe Chen, Baolong Bi, and Xueqi Cheng. 2025.
\newblock Innate reasoning is not enough: In-context learning enhances reasoning large language models with less overthinking.
\newblock \emph{arXiv preprint arXiv:2503.19602}.

\bibitem[{Guerreiro et~al.(2023)Guerreiro, Voita, and Martins}]{guerreiro-etal-2023-looking}
Nuno~M. Guerreiro, Elena Voita, and Andr{\'e} Martins. 2023.
\newblock \href {https://doi.org/10.18653/v1/2023.eacl-main.75} {Looking for a needle in a haystack: A comprehensive study of hallucinations in neural machine translation}.
\newblock In \emph{Proceedings of the 17th Conference of the European Chapter of the Association for Computational Linguistics}, pages 1059--1075, Dubrovnik, Croatia. Association for Computational Linguistics.

\bibitem[{Guo et~al.(2017)Guo, Pleiss, Sun, and Weinberger}]{guo2017calibration}
Chuan Guo, Geoff Pleiss, Yu~Sun, and Kilian~Q Weinberger. 2017.
\newblock On calibration of modern neural networks.
\newblock In \emph{International conference on machine learning}, pages 1321--1330. PMLR.

\bibitem[{Huang et~al.(2023)Huang, Song, Wang, Zhao, Chen, Juefei-Xu, and Ma}]{huang2023look}
Yuheng Huang, Jiayang Song, Zhijie Wang, Shengming Zhao, Huaming Chen, Felix Juefei-Xu, and Lei Ma. 2023.
\newblock Look before you leap: An exploratory study of uncertainty measurement for large language models.
\newblock \emph{arXiv preprint arXiv:2307.10236}.

\bibitem[{Izacard and Grave(2021)}]{izacard2021leveraging}
Gautier Izacard and Edouard Grave. 2021.
\newblock \href {https://doi.org/10.18653/v1/2021.eacl-main.74} {Leveraging passage retrieval with generative models for open domain question answering}.
\newblock In \emph{Proceedings of the 16th Conference of the European Chapter of the Association for Computational Linguistics: Main Volume}, pages 874--880, Online. Association for Computational Linguistics.

\bibitem[{Jeong et~al.(2024)Jeong, Baek, Cho, Hwang, and Park}]{jeong2024adaptive}
Soyeong Jeong, Jinheon Baek, Sukmin Cho, Sung~Ju Hwang, and Jong~C Park. 2024.
\newblock \href {https://arxiv.org/abs/2403.14403} {Adaptive-rag: Learning to adapt retrieval-augmented large language models through question complexity}.
\newblock \emph{ArXiv preprint}, abs/2403.14403.

\bibitem[{Jiang et~al.(2024)Jiang, Sablayrolles, Roux, Mensch, Savary, Bamford, Chaplot, Casas, Hanna, Bressand et~al.}]{jiang2024mixtral}
Albert~Q Jiang, Alexandre Sablayrolles, Antoine Roux, Arthur Mensch, Blanche Savary, Chris Bamford, Devendra~Singh Chaplot, Diego de~las Casas, Emma~Bou Hanna, Florian Bressand, et~al. 2024.
\newblock \href {https://arxiv.org/abs/2401.04088} {Mixtral of experts}.
\newblock \emph{ArXiv preprint}, abs/2401.04088.

\bibitem[{Jiang et~al.(2023)Jiang, Xu, Gao, Sun, Liu, Dwivedi-Yu, Yang, Callan, and Neubig}]{jiang2023active}
Zhengbao Jiang, Frank~F Xu, Luyu Gao, Zhiqing Sun, Qian Liu, Jane Dwivedi-Yu, Yiming Yang, Jamie Callan, and Graham Neubig. 2023.
\newblock Active retrieval augmented generation.
\newblock In \emph{Proceedings of the 2023 Conference on Empirical Methods in Natural Language Processing}, pages 7969--7992.

\bibitem[{Jin et~al.(2019)Jin, Dhingra, Liu, Cohen, and Lu}]{jin2019pubmedqa}
Qiao Jin, Bhuwan Dhingra, Zhengping Liu, William Cohen, and Xinghua Lu. 2019.
\newblock \href {https://doi.org/10.18653/v1/D19-1259} {{P}ub{M}ed{QA}: A dataset for biomedical research question answering}.
\newblock In \emph{Proceedings of the 2019 Conference on Empirical Methods in Natural Language Processing and the 9th International Joint Conference on Natural Language Processing (EMNLP-IJCNLP)}, pages 2567--2577, Hong Kong, China. Association for Computational Linguistics.

\bibitem[{Joshi et~al.(2017{\natexlab{a}})Joshi, Choi, Weld, and Zettlemoyer}]{triviaqa}
Mandar Joshi, Eunsol Choi, Daniel Weld, and Luke Zettlemoyer. 2017{\natexlab{a}}.
\newblock \href {https://doi.org/10.18653/v1/P17-1147} {{T}rivia{QA}: A large scale distantly supervised challenge dataset for reading comprehension}.
\newblock In \emph{Proceedings of the 55th Annual Meeting of the Association for Computational Linguistics (Volume 1: Long Papers)}, pages 1601--1611, Vancouver, Canada. Association for Computational Linguistics.

\bibitem[{Joshi et~al.(2017{\natexlab{b}})Joshi, Choi, Weld, and Zettlemoyer}]{joshi2017triviaqa}
Mandar Joshi, Eunsol Choi, Daniel~S Weld, and Luke Zettlemoyer. 2017{\natexlab{b}}.
\newblock Triviaqa: A large scale distantly supervised challenge dataset for reading comprehension.
\newblock In \emph{Proceedings of the 55th Annual Meeting of the Association for Computational Linguistics (Volume 1: Long Papers)}, pages 1601--1611.

\bibitem[{Kadavath et~al.(2022)Kadavath, Conerly, Askell, Henighan, Drain, Perez, Schiefer, Hatfield-Dodds, DasSarma, Tran-Johnson et~al.}]{kadavath2022language}
Saurav Kadavath, Tom Conerly, Amanda Askell, Tom Henighan, Dawn Drain, Ethan Perez, Nicholas Schiefer, Zac Hatfield-Dodds, Nova DasSarma, Eli Tran-Johnson, et~al. 2022.
\newblock Language models (mostly) know what they know.
\newblock \emph{arXiv preprint arXiv:2207.05221}.

\bibitem[{Karpukhin et~al.(2020)Karpukhin, O{\u{g}}uz, Min, Lewis, Wu, Edunov, Chen, and Yih}]{karpukhin2020dense}
Vladimir Karpukhin, Barlas O{\u{g}}uz, Sewon Min, Patrick Lewis, Ledell Wu, Sergey Edunov, Danqi Chen, and Wen-tau Yih. 2020.
\newblock Dense passage retrieval for open-domain question answering.
\newblock \emph{arXiv preprint arXiv:2004.04906}.

\bibitem[{Kwiatkowski et~al.(2019{\natexlab{a}})Kwiatkowski, Palomaki, Redfield, Collins, Parikh, Alberti, Epstein, Polosukhin, Devlin, Lee, Toutanova, Jones, Kelcey, Chang, Dai, Uszkoreit, Le, and Petrov}]{nq}
Tom Kwiatkowski, Jennimaria Palomaki, Olivia Redfield, Michael Collins, Ankur Parikh, Chris Alberti, Danielle Epstein, Illia Polosukhin, Jacob Devlin, Kenton Lee, Kristina Toutanova, Llion Jones, Matthew Kelcey, Ming-Wei Chang, Andrew~M. Dai, Jakob Uszkoreit, Quoc Le, and Slav Petrov. 2019{\natexlab{a}}.
\newblock \href {https://doi.org/10.1162/tacl_a_00276} {Natural questions: A benchmark for question answering research}.
\newblock \emph{Transactions of the Association for Computational Linguistics}, 7:452--466.

\bibitem[{Kwiatkowski et~al.(2019{\natexlab{b}})Kwiatkowski, Palomaki, Redfield, Collins, Parikh, Alberti, Epstein, Polosukhin, Devlin, Lee et~al.}]{kwiatkowski2019natural}
Tom Kwiatkowski, Jennimaria Palomaki, Olivia Redfield, Michael Collins, Ankur Parikh, Chris Alberti, Danielle Epstein, Illia Polosukhin, Jacob Devlin, Kenton Lee, et~al. 2019{\natexlab{b}}.
\newblock Natural questions: a benchmark for question answering research.
\newblock \emph{Transactions of the Association for Computational Linguistics}, 7:453--466.

\bibitem[{Lewis et~al.(2020)Lewis, Perez, Piktus, Petroni, Karpukhin, Goyal, K{\"{u}}ttler, Lewis, Yih, Rockt{\"{a}}schel, Riedel, and Kiela}]{lewis2020retrieval}
Patrick S.~H. Lewis, Ethan Perez, Aleksandra Piktus, Fabio Petroni, Vladimir Karpukhin, Naman Goyal, Heinrich K{\"{u}}ttler, Mike Lewis, Wen{-}tau Yih, Tim Rockt{\"{a}}schel, Sebastian Riedel, and Douwe Kiela. 2020.
\newblock \href {https://proceedings.neurips.cc/paper/2020/hash/6b493230205f780e1bc26945df7481e5-Abstract.html} {Retrieval-augmented generation for knowledge-intensive {NLP} tasks}.
\newblock In \emph{Advances in Neural Information Processing Systems 33: Annual Conference on Neural Information Processing Systems 2020, NeurIPS 2020, December 6-12, 2020, virtual}.

\bibitem[{Li et~al.(2024)Li, Yuan, and Zhang}]{li2024enhancing}
Jiarui Li, Ye~Yuan, and Zehua Zhang. 2024.
\newblock Enhancing llm factual accuracy with rag to counter hallucinations: A case study on domain-specific queries in private knowledge-bases.
\newblock \emph{arXiv preprint arXiv:2403.10446}.

\bibitem[{Lin et~al.(2022)Lin, Hilton, and Evans}]{lin2022teaching}
Stephanie Lin, Jacob Hilton, and Owain Evans. 2022.
\newblock Teaching models to express their uncertainty in words.
\newblock \emph{Transactions on Machine Learning Research}.

\bibitem[{Liu et~al.(2024{\natexlab{a}})Liu, Lin, Hewitt, Paranjape, Bevilacqua, Petroni, and Liang}]{liu2024lost}
Nelson~F Liu, Kevin Lin, John Hewitt, Ashwin Paranjape, Michele Bevilacqua, Fabio Petroni, and Percy Liang. 2024{\natexlab{a}}.
\newblock Lost in the middle: How language models use long contexts.
\newblock \emph{Transactions of the Association for Computational Linguistics}, 12:157--173.

\bibitem[{Liu et~al.(2024{\natexlab{b}})Liu, Ping, Roy, Xu, Shoeybi, and Catanzaro}]{liu2024chatqa}
Zihan Liu, Wei Ping, Rajarshi Roy, Peng Xu, Mohammad Shoeybi, and Bryan Catanzaro. 2024{\natexlab{b}}.
\newblock \href {https://arxiv.org/abs/2401.10225} {Chatqa: Surpassing gpt-4 on conversational qa and rag}.
\newblock In \emph{NeurIPS}.

\bibitem[{Meta-AI(2024)}]{llama3}
Meta-AI. 2024.
\newblock Llama 3 model card.

\bibitem[{Ni et~al.(2024)Ni, Bi, Guo, and Cheng}]{ni2024llms}
Shiyu Ni, Keping Bi, Jiafeng Guo, and Xueqi Cheng. 2024.
\newblock When do llms need retrieval augmentation? mitigating llms' overconfidence helps retrieval augmentation.
\newblock \emph{arXiv preprint arXiv:2402.11457}.

\bibitem[{Ni et~al.(2025)Ni, Bi, Guo, Yu, Bi, and Cheng}]{ni2025towards}
Shiyu Ni, Keping Bi, Jiafeng Guo, Lulu Yu, Baolong Bi, and Xueqi Cheng. 2025.
\newblock Towards fully exploiting llm internal states to enhance knowledge boundary perception.
\newblock \emph{arXiv preprint arXiv:2502.11677}.

\bibitem[{OpenAI(2022)}]{chatgpt}
OpenAI. 2022.
\newblock Introducing {ChatGPT}.

\bibitem[{Pasca(2019)}]{pasca-2019-wikipedia}
Marius Pasca. 2019.
\newblock \href {https://doi.org/10.18653/v1/P19-4005} {{W}ikipedia as a resource for text analysis and retrieval}.
\newblock In \emph{Proceedings of the 57th Annual Meeting of the Association for Computational Linguistics: Tutorial Abstracts}, page~24, Florence, Italy. Association for Computational Linguistics.

\bibitem[{Radford et~al.(2019)Radford, Wu, Child, Luan, Amodei, Sutskever et~al.}]{radford2019language}
Alec Radford, Jeffrey Wu, Rewon Child, David Luan, Dario Amodei, Ilya Sutskever, et~al. 2019.
\newblock Language models are unsupervised multitask learners.
\newblock \emph{OpenAI blog}, 1(8):9.

\bibitem[{Rafailov et~al.(2024)Rafailov, Sharma, Mitchell, Manning, Ermon, and Finn}]{rafailov2024direct}
Rafael Rafailov, Archit Sharma, Eric Mitchell, Christopher~D Manning, Stefano Ermon, and Chelsea Finn. 2024.
\newblock Direct preference optimization: Your language model is secretly a reward model.
\newblock \emph{Advances in Neural Information Processing Systems}, 36.

\bibitem[{Raja et~al.(2024)Raja, Yuvaraajan et~al.}]{raja2024rag}
Mahimai Raja, E~Yuvaraajan, et~al. 2024.
\newblock A rag-based medical assistant especially for infectious diseases.
\newblock In \emph{2024 International Conference on Inventive Computation Technologies (ICICT)}, pages 1128--1133. IEEE.

\bibitem[{Reji et~al.(2024)Reji, Sheik, Sharon, Rai, and Nirmala}]{reji2024enhancing}
Sneha~Ann Reji, Reshma Sheik, A~Sharon, Avisha Rai, and S~Jaya Nirmala. 2024.
\newblock Enhancing llm performance on legal textual entailment with few-shot cot-based rag.
\newblock In \emph{2024 IEEE International Conference on Signal Processing, Informatics, Communication and Energy Systems (SPICES)}, pages 1--6. IEEE.

\bibitem[{Shuster et~al.(2021)Shuster, Poff, Chen, Kiela, and Weston}]{shuster-etal-2021-retrieval-augmentation}
Kurt Shuster, Spencer Poff, Moya Chen, Douwe Kiela, and Jason Weston. 2021.
\newblock \href {https://doi.org/10.18653/v1/2021.findings-emnlp.320} {Retrieval augmentation reduces hallucination in conversation}.
\newblock In \emph{Findings of the Association for Computational Linguistics: EMNLP 2021}, pages 3784--3803, Punta Cana, Dominican Republic. Association for Computational Linguistics.

\bibitem[{Song et~al.(2024)Song, Sim, Bhardwaj, Chieu, Majumder, and Poria}]{song2024measuring}
Maojia Song, Shang~Hong Sim, Rishabh Bhardwaj, Hai~Leong Chieu, Navonil Majumder, and Soujanya Poria. 2024.
\newblock Measuring and enhancing trustworthiness of llms in rag through grounded attributions and learning to refuse.
\newblock \emph{arXiv preprint arXiv:2409.11242}.

\bibitem[{Stengel-Eskin et~al.(2024)Stengel-Eskin, Hase, and Bansal}]{stengel2024lacie}
Elias Stengel-Eskin, Peter Hase, and Mohit Bansal. 2024.
\newblock Lacie: Listener-aware finetuning for confidence calibration in large language models.
\newblock \emph{arXiv preprint arXiv:2405.21028}.

\bibitem[{Tan et~al.(2024)Tan, Sun, Yang, Wang, Cao, and Cheng}]{tan2024blinded}
Hexiang Tan, Fei Sun, Wanli Yang, Yuanzhuo Wang, Qi~Cao, and Xueqi Cheng. 2024.
\newblock Blinded by generated contexts: How language models merge generated and retrieved contexts for open-domain qa?
\newblock \emph{arXiv preprint arXiv:2401.11911}.

\bibitem[{Thakur et~al.(2024)Thakur, Bonifacio, Zhang, Ogundepo, Kamalloo, Alfonso-Hermelo, Li, Liu, Chen, Rezagholizadeh, and Lin}]{thakur-etal-2024-knowing}
Nandan Thakur, Luiz Bonifacio, Crystina Zhang, Odunayo Ogundepo, Ehsan Kamalloo, David Alfonso-Hermelo, Xiaoguang Li, Qun Liu, Boxing Chen, Mehdi Rezagholizadeh, and Jimmy Lin. 2024.
\newblock \href {https://doi.org/10.18653/v1/2024.findings-emnlp.730} {{``}knowing when you don{'}t know{''}: A multilingual relevance assessment dataset for robust retrieval-augmented generation}.
\newblock In \emph{Findings of the Association for Computational Linguistics: EMNLP 2024}, pages 12508--12526, Miami, Florida, USA. Association for Computational Linguistics.

\bibitem[{Tian et~al.(2023)Tian, Mitchell, Zhou, Sharma, Rafailov, Yao, Finn, and Manning}]{tian2023just}
Katherine Tian, Eric Mitchell, Allan Zhou, Archit Sharma, Rafael Rafailov, Huaxiu Yao, Chelsea Finn, and Christopher~D Manning. 2023.
\newblock Just ask for calibration: Strategies for eliciting calibrated confidence scores from language models fine-tuned with human feedback.
\newblock In \emph{Proceedings of the 2023 Conference on Empirical Methods in Natural Language Processing}, pages 5433--5442.

\bibitem[{Touvron et~al.(2023)Touvron, Martin, Stone, Albert, Almahairi, Babaei, Bashlykov, Batra, Bhargava, Bhosale et~al.}]{touvron2023llama2}
Hugo Touvron, Louis Martin, Kevin Stone, Peter Albert, Amjad Almahairi, Yasmine Babaei, Nikolay Bashlykov, Soumya Batra, Prajjwal Bhargava, Shruti Bhosale, et~al. 2023.
\newblock \href {https://arxiv.org/abs/2307.09288} {Llama 2: Open foundation and fine-tuned chat models}.
\newblock \emph{ArXiv preprint}, abs/2307.09288.

\bibitem[{Varshney et~al.(2023)Varshney, Yao, Zhang, Chen, and Yu}]{varshney2023stitch}
Neeraj Varshney, Wenlin Yao, Hongming Zhang, Jianshu Chen, and Dong Yu. 2023.
\newblock A stitch in time saves nine: Detecting and mitigating hallucinations of llms by validating low-confidence generation.
\newblock \emph{arXiv preprint arXiv:2307.03987}.

\bibitem[{Wang et~al.(2024)Wang, Cheng, Guo, Yue, Ding, Xu, Wang, Hu, Zhang, and Zhang}]{wang2024evaluating}
Cunxiang Wang, Sirui Cheng, Qipeng Guo, Yuanhao Yue, Bowen Ding, Zhikun Xu, Yidong Wang, Xiangkun Hu, Zheng Zhang, and Yue Zhang. 2024.
\newblock Evaluating open-qa evaluation.
\newblock \emph{Advances in Neural Information Processing Systems}, 36.

\bibitem[{Wang et~al.(2022)Wang, Wei, Schuurmans, Le, Chi, Narang, Chowdhery, and Zhou}]{wang2022self}
Xuezhi Wang, Jason Wei, Dale Schuurmans, Quoc Le, Ed~Chi, Sharan Narang, Aakanksha Chowdhery, and Denny Zhou. 2022.
\newblock Self-consistency improves chain of thought reasoning in language models.
\newblock \emph{arXiv preprint arXiv:2203.11171}.

\bibitem[{Wei et~al.(2022)Wei, Wang, Schuurmans, Bosma, Xia, Chi, Le, Zhou et~al.}]{wei2022chain}
Jason Wei, Xuezhi Wang, Dale Schuurmans, Maarten Bosma, Fei Xia, Ed~Chi, Quoc~V Le, Denny Zhou, et~al. 2022.
\newblock Chain-of-thought prompting elicits reasoning in large language models.
\newblock \emph{Advances in neural information processing systems}, 35:24824--24837.

\bibitem[{Xiong et~al.(2024)Xiong, Hu, Lu, LI, Fu, He, and Hooi}]{xiong2024can}
Miao Xiong, Zhiyuan Hu, Xinyang Lu, YIFEI LI, Jie Fu, Junxian He, and Bryan Hooi. 2024.
\newblock Can llms express their uncertainty? an empirical evaluation of confidence elicitation in llms.
\newblock In \emph{The Twelfth International Conference on Learning Representations}.

\bibitem[{Xu et~al.(2024{\natexlab{a}})Xu, Zhu, Ma, Zhang, Fan, Chen, and Yu}]{xu2024rejection}
Hongshen Xu, Zichen Zhu, Da~Ma, Situo Zhang, Shuai Fan, Lu~Chen, and Kai Yu. 2024{\natexlab{a}}.
\newblock Rejection improves reliability: Training llms to refuse unknown questions using rl from knowledge feedback.
\newblock \emph{arXiv preprint arXiv:2403.18349}.

\bibitem[{Xu et~al.(2024{\natexlab{b}})Xu, Fei, Pan, Liu, Lee, and Hsu}]{xu2024faithful}
Jundong Xu, Hao Fei, Liangming Pan, Qian Liu, Mong-Li Lee, and Wynne Hsu. 2024{\natexlab{b}}.
\newblock Faithful logical reasoning via symbolic chain-of-thought.
\newblock \emph{arXiv preprint arXiv:2405.18357}.

\bibitem[{Yang et~al.(2023)Yang, Chern, Qiu, Neubig, and Liu}]{yang2023alignment}
Yuqing Yang, Ethan Chern, Xipeng Qiu, Graham Neubig, and Pengfei Liu. 2023.
\newblock Alignment for honesty.
\newblock \emph{arXiv preprint arXiv:2312.07000}.

\bibitem[{Yepes et~al.(2024)Yepes, You, Milczek, Laverde, and Li}]{yepes2024financial}
Antonio~Jimeno Yepes, Yao You, Jan Milczek, Sebastian Laverde, and Renyu Li. 2024.
\newblock Financial report chunking for effective retrieval augmented generation.
\newblock \emph{arXiv preprint arXiv:2402.05131}.

\bibitem[{Yoran et~al.(2024)Yoran, Wolfson, Ram, and Berant}]{robustlm}
Ori Yoran, Tomer Wolfson, Ori Ram, and Jonathan Berant. 2024.
\newblock \href {https://openreview.net/forum?id=ZS4m74kZpH} {Making retrieval-augmented language models robust to irrelevant context}.
\newblock In \emph{ICLR}.

\bibitem[{Zhang et~al.(2024{\natexlab{a}})Zhang, Diao, Lin, Fung, Lian, Wang, Chen, Ji, and Zhang}]{zhang2024r}
Hanning Zhang, Shizhe Diao, Yong Lin, Yi~Fung, Qing Lian, Xingyao Wang, Yangyi Chen, Heng Ji, and Tong Zhang. 2024{\natexlab{a}}.
\newblock R-tuning: Instructing large language models to say ‘i don’t know’.
\newblock In \emph{Proceedings of the 2024 Conference of the North American Chapter of the Association for Computational Linguistics: Human Language Technologies (Volume 1: Long Papers)}, pages 7106--7132.

\bibitem[{Zhang et~al.(2024{\natexlab{b}})Zhang, Patil, Jain, Shen, Zaharia, Stoica, and Gonzalez}]{zhang2024raft}
Tianjun Zhang, Shishir~G Patil, Naman Jain, Sheng Shen, Matei Zaharia, Ion Stoica, and Joseph~E. Gonzalez. 2024{\natexlab{b}}.
\newblock \href {https://openreview.net/forum?id=rzQGHXNReU} {{RAFT}: Adapting language model to domain specific {RAG}}.
\newblock In \emph{COLM}.

\bibitem[{Zhao et~al.(2024)Zhao, Zhang, Pan, Yao, Yu, Wu, and Chen}]{zhao2024fact}
Xinran Zhao, Hongming Zhang, Xiaoman Pan, Wenlin Yao, Dong Yu, Tongshuang Wu, and Jianshu Chen. 2024.
\newblock Fact-and-reflection ({F}a{R}) improves confidence calibration of large language models.
\newblock In \emph{Findings of the Association for Computational Linguistics ACL 2024}, pages 8702--8718.

\bibitem[{Zhou et~al.(2023{\natexlab{a}})Zhou, Sch{\"a}rli, Hou, Wei, Scales, Wang, Schuurmans, Cui, Bousquet, Le, and Chi}]{zhou2023leasttomost}
Denny Zhou, Nathanael Sch{\"a}rli, Le~Hou, Jason Wei, Nathan Scales, Xuezhi Wang, Dale Schuurmans, Claire Cui, Olivier Bousquet, Quoc~V Le, and Ed~H. Chi. 2023{\natexlab{a}}.
\newblock Least-to-most prompting enables complex reasoning in large language models.
\newblock In \emph{The Eleventh International Conference on Learning Representations}.

\bibitem[{Zhou et~al.(2023{\natexlab{b}})Zhou, Zhang, Poon, and Chen}]{zhou-etal-2023-context}
Wenxuan Zhou, Sheng Zhang, Hoifung Poon, and Muhao Chen. 2023{\natexlab{b}}.
\newblock \href {https://doi.org/10.18653/v1/2023.findings-emnlp.968} {Context-faithful prompting for large language models}.
\newblock In \emph{Findings of the Association for Computational Linguistics: EMNLP 2023}, pages 14544--14556, Singapore. Association for Computational Linguistics.

\end{thebibliography}
\appendix

\section{The Details of the Knowledge Quadrants}
\label{appendix:knowledge_quadrants}

\bluecheck\greencheck\ represents the most ideal but trivial scenario, where both the model's parametric knowledge and retrieved passages contain the correct information.

\bluecheck\greencross\ occurs when $q \in \KBparam$ but $q \notin \KBr$, indicating that while the model has the necessary parametric knowledge, the retriever fails to find relevant passages. In such cases, retrieval is unnecessary and the model should rely on its parametric knowledge. Many adaptive RAG methods \cite{jeong2024adaptive,asai2024selfrag} focus on identifying and handling this scenario.

\bluecross\greencheck\ represents the core scenario that RAG systems are designed to handle, where $q \in \KBr$ but $q \notin \KBparam$. Here, while the model lacks the necessary parametric knowledge, the retrieved passages contain the correct information. However, even with the correct information present in the retrieved passages, the model may fail to utilize it effectively due to issues such as "lost in the middle" \cite{liu2024lost}.

RAFT acctually enhances the RAG system's answer accuracy across both \bluecheck\greencross\ and \bluecross\greencheck\ scenarios by addressing their distinct challenges: For \bluecheck\greencross: RAFT teaches the model to rely on its parametric knowledge when retrieved passages are noisy.
For \bluecross\greencheck: RAFT helps the model better utilize information from retrieved passages. So the RAFT get some emprical success in a some previous work \cite{fang2024enhancing,robustlm,zhang2024raft,liu2024chatqa}.

In the \bluecross\greencross\ case ($q \notin \KBparam \cup \KBr$), neither the model's parametric knowledge nor the retrieved passages contain the correct information. In such case, the model should ideally abstain from answering to maintain faithfulness and avoid hallucination. However, current RAFT-trained models are conditioned to always generate an answer, even when the query is out of $\KBrag$. This leads to an overly aggressive response pattern that prioritizes answer generation over honesty, potentially producing misleading or entirely fabricated responses. While RAFT approaches may improve surface-level metrics like answer accuracy, it fundamentally compromises the system's reliability and trustworthiness. In this work, we specifically focus on addressing this critical gap by developing methods that enable models to recognize when a query falls outside of $\KBrag$ and appropriately respond with "I don't know". This capability is essential for deploying RAG systems in high-stakes applications where the cost of hallucination and misinformation can be severe.

\section{Related works}
\subsection{Retrieval-Augmented Generation } RAG \cite{lewis2020retrieval,borgeaud2022improving,izacard2021leveraging,zhang2024raft} is a widely adopted paradigm for augmenting large language models (LLMs) with external knowledge. By integrating a retrieval system, RAG enables models to access and utilize external knowledge sources during generation, overcoming the limitations of static, parameterized knowledge in LLMs. This approach has shown significant promise in tasks requiring factual accuracy, domain-specific knowledge \cite{zhang2024raft}, and up-to-date information \cite{li2024enhancing}.  Despite its advantages, the effectiveness of RAG heavily depends on the quality of the retrieved passages. Current retrieval systems often fail to guarantee complete relevance, introducing noisy contexts into the generation process. To address this challenge, Retrieval-Augmented Fine-Tuning (RAFT) \cite{zhang2024raft, fang2024enhancing, liu2024chatqa} has been proposed. RAFT fine-tunes models with a mixture of retrieved contexts, including both clean and noisy passages, encouraging robustness to imperfect retrieval results. 

However, RAFT-trained models exhibit a critical limitation: they are conditioned to answer queries even when provided with entirely noisy  contexts. This over-reliance on retrieved information increases the risk of generating hallucinated or misleading responses, especially in high-stakes applications. Our work builds on this understanding by addressing the overlooked issue of enabling RAFT-trained models to acknowledge uncertainty and respond with “I don't know” when appropriate.

\subsection{Honest Alignment in Large Language Models}
Honesty is a foundational principle in aligning large language models (LLMs) with human values. It requires models to accurately express their knowledge, recognize their limitations, and avoid misleading users when uncertain. Honesty encompasses two critical components: self-knowledge and self-expression. Self-Knowledge refers to the model's ability to discern what it knows and doesn't know, enabling it to explicitly admit uncertainty (e.g., responding “I don't know”) when necessary. This capability is crucial for mitigating hallucinations and ensuring model reliability in high-stakes applications. Current methods to improve self-knowledge include: Training-free approaches: These leverage predictive probabilities \cite{duan2024shifting}, prompting strategies 
 \cite{zhou2023leasttomost,kadavath2022language,zhao2024fact} (e.g., Chain-of-Thought reasoning), and sampling/aggregation 
techniques to elicit calibrated confidence from models \cite{tian2023just,guo2017calibration,xiong2024can}. While effective in some contexts, these approaches often struggle with free-form generation and require significant computational overhead. Training-based approaches: Methods such as supervised fine-tuning and reinforcement learning aim to teach models to abstain from answering uncertain queries or provide confidence scores alongside responses 
 \cite{yang2023alignment, zhang2024r,jiang2024mixtral,zhou2023leasttomost,gao2024best,xu2024faithful,stengel2024lacie}. However, these works only consider the LLM's parametric knowledge boundary, and ignore the knowledge boundary of the retrieval system.

%

Our work builds on these foundations, endowing the retrieval-augmented models with the ability to acknowledge uncertainty under noisy contexts based on the preference training on the four knowledge  quadrants.

\textbf{Comparison with the existing works: }
Most of the current raft work \cite{robustlm,fang2024enhancing,liu2024chatqa} and rag work \cite{asai2023self, lewis2020retrieval} try to improve the model's ability on the accuracy of response and ignore the faithfulness of the response. And we have shown that the success of the current raft work is built on the sacrifice of the faithfulness of the response.  The model actually becomes an aggressively omniscient model. \citet{cheng2024can,feng-etal-2024-dont,xu2024rejection} align the model to abstain when the model can not answer the query. These work actually only focus on the knowledge boundary of the LLM itself. But in the RAG scenario, the knowledge boundary is actually the combination of the LLM knowledge boundary and the retrieval knowledge boundary. \citet{song2024measuring, thakur-etal-2024-knowing} align the  model to refuse answer when the retrieved passages are noisy. But they ignore the knowledge boundary of the LLM itself. \textbf{Our work is the first work that simultaneously considers the knowledge boundary of the LLM itself and the retrieval knowledge boundary} and aligns the model to refuse answer only when the query is out of the both knowledge boundaries.

\section{Baseline Methods}
\label{appendix:baselines}

We compare our approach against several state-of-the-art baselines and corresponding Llama family base models.

\paragraph{Base Models:}
\begin{itemize}
    \item \textbf{Llama2-7B} \cite{touvron2023llama2}: A member of Llama2 family with 7 billion parameters, which is released in July 2023.
    \item \textbf{Llama2-13B} \cite{touvron2023llama2}: A member of Llama2 family with 13 billion parameters, which is released in July 2023.
    \item \textbf{Llama3-8B} \cite{llama3}: A member of Llama3 family with 8 billion parameters, which is released in April 2024.
\end{itemize}

\paragraph{RAFT Models:}
\begin{itemize}
  \item \textbf{RAAT} \cite{fang2024enhancing}: A model that employs adaptive adversarial training to handle three types of retrieval noises (relevant, irrelevant, and counterfactual). During training, it dynamically selects the most challenging noise type based on the model's current performance and uses multi-task learning to enhance noise awareness.
  \item \textbf{Ret-Robust} \cite{robustlm}: A model that trains with a mixture of relevant and irrelevant retrieved contexts. For each training example, it retrieves either top-1, low-ranked, or random passages with equal probability to teach the model when to use or ignore retrieved information.
  \item \textbf{ChatQA} \cite{liu2024chatqa}: A two-stage instruction tuning approach that outperforms GPT-4 on retrieval-augmented generation and conversational QA tasks. It first performs supervised fine-tuning to enhance basic instruction following capabilities, then conducts context-enhanced instruction tuning specifically for dialogue QA and RAG tasks.
\end{itemize}

\paragraph{Calibration Methods:}
These methods use post-hoc techniques to predict whether the retrieved passages are relevant to the question or if the model is likely to hallucinate, which can trigger a refusal to answer.
\begin{itemize}
  \item \textbf{P(True)} \cite{kadavath2022language}: Uses prompt-based evaluation to assess the correctness of model generations, leveraging the observation that LLMs are relatively well-calibrated in self-evaluation tasks.
  \item \textbf{Logits}: Implements various methods from previous studies \cite{guerreiro-etal-2023-looking, kadavath2022language, varshney2023stitch, huang2023look} that aggregate output token probabilities or logits to score LLM confidence for error detection.
\end{itemize}

\begin{figure*}[t]
    \centering
    \includegraphics[width=\linewidth]{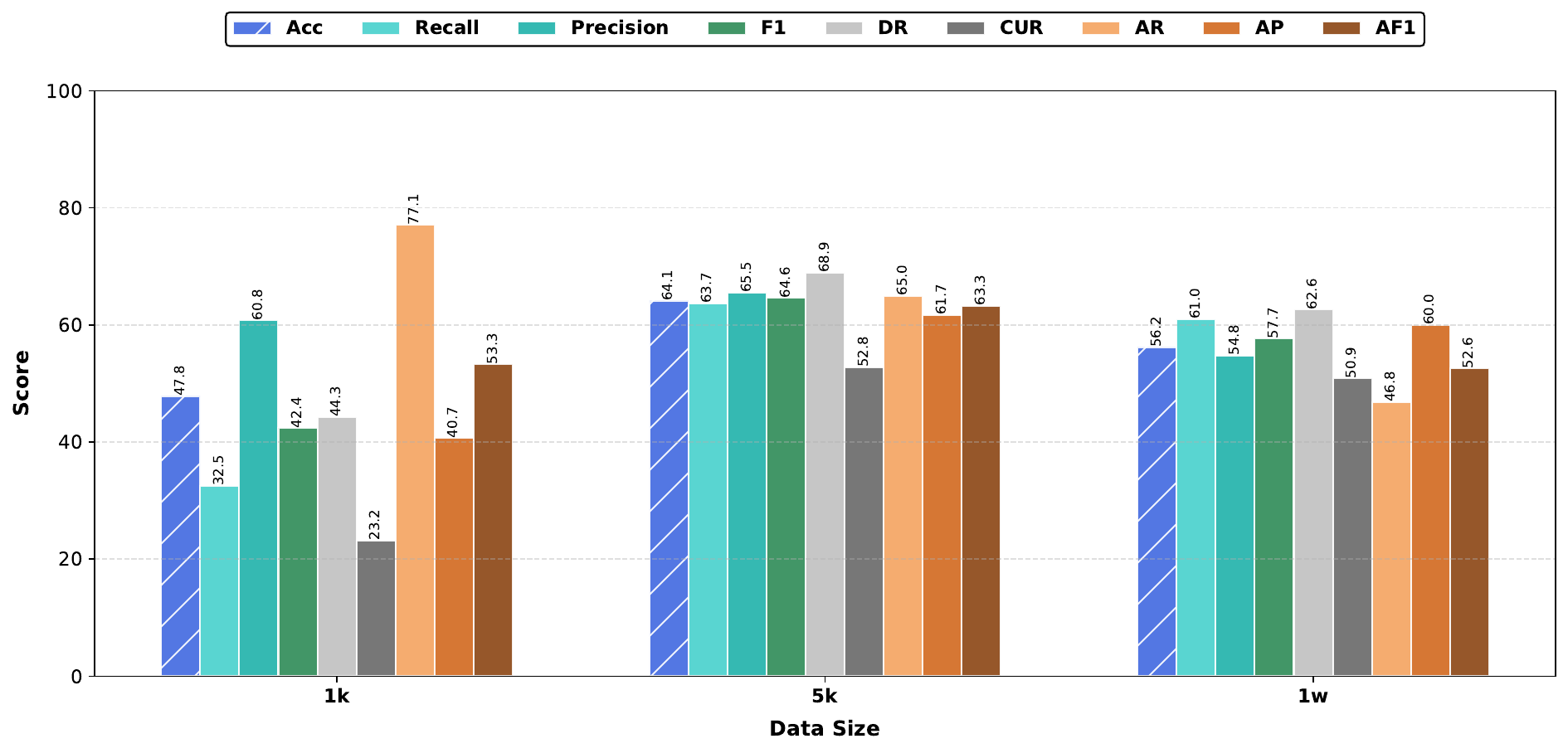}
    \caption{Experiments across DPO data size. (IDK ratio=0.7, loss weights $\beta$=1.0, $\gamma$=0.5)}
    \label{fig:hyper-datasize}
\end{figure*}

We also include two widely-used baseline approaches:
\begin{itemize}
  \item \textbf{ICL}: We implement in-context learning using a prompt template with three carefully curated demonstration examples: one showing appropriate abstention for out-of-knowledge-boundary queries, and two showcasing correct answer generation for in-boundary queries. This balanced demonstration set helps the model learn both when to answer and when to abstain.
  \item \textbf{Consistency} \cite{wang2022self}: Uses the consistency of the model's responses to determine whether it should abstain from answering.
\end{itemize}

\section{Hyper-parameter experiments}
\label{appendix:hyperparameter}
\begin{figure}[h]
    \centering
    \includegraphics[width=\linewidth]{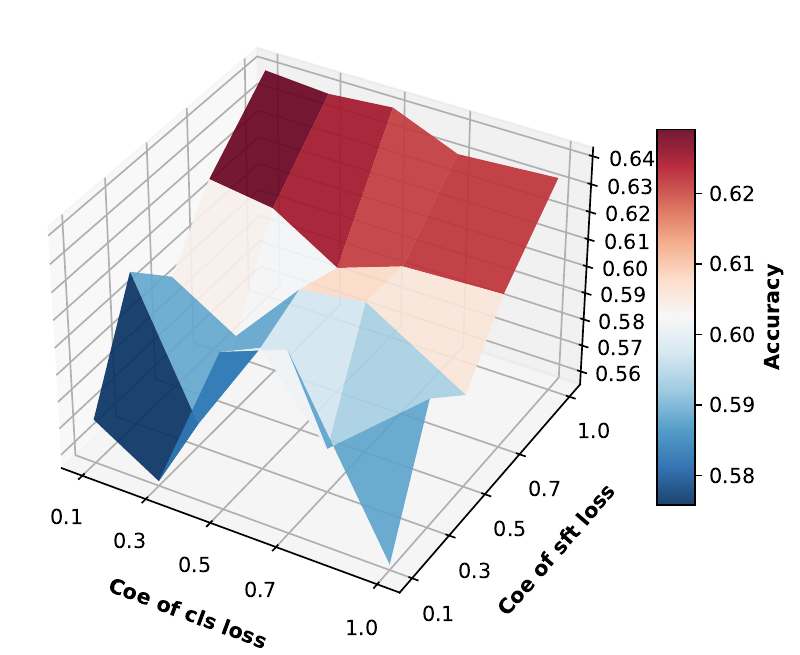}
    \caption{Experiments across multi-objective loss weights. (DPO data size=5k, IDK ratio=0.7)}
    \label{fig:hyper-coeloss}
\end{figure}

\paragraph{Multi-Objective Loss} Adjusting the weights of the multi-objective loss significantly impacts model's overall quality. As shown in Figure~\ref{fig:hyper-coeloss}, increasing the weight of the SFT loss generally leads to steady improvements, which is in line with our hypothesis. The experiments confirm that SFT effectively assists in aligning with the chosen data, demonstrating strong auxiliary alignment effects. Meanwhile, the classification loss (CLS) is not without merit; it plays a critical role when combined with the SFT loss, achieving optimal performance within the weight range of 0.5 to 0.7. This highlights the synergistic interplay between the two loss components under balanced configurations.

\begin{figure*}[h]
    \centering
    \includegraphics[width=\linewidth]{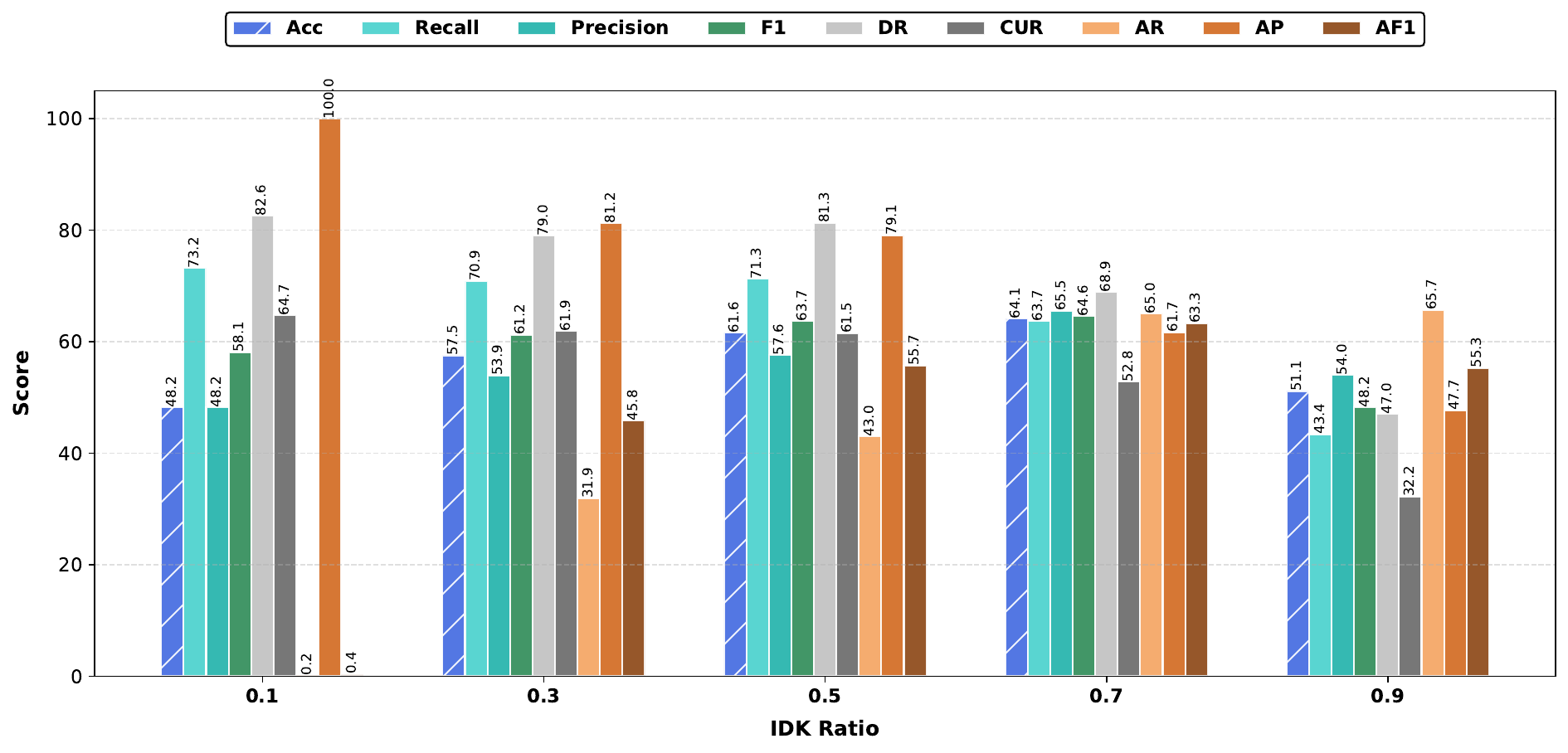}
    \caption{Experiments across IDK ratio. (DPO data size=5k, loss weights $\beta$=1.0, $\gamma$=0.5)}
    \label{fig:hyper-idkratio}
\end{figure*}

\paragraph{Data Size} Statistics in Figure \ref{fig:hyper-datasize} show that 5k DPO preference data achieves competitive performance in terms of overall quality(OQ Acc), answer quality(AQ F1), and abstain quality(AbQ F1). Reducing data to 20\% sharply degrades the outcomes, which indicates the significance of sufficient training data. However, when data size grows to 10k, it seems increased noise-potentially introduced by scaling without rigorous quality control-lead to performance degradation. This pattern emphasizes the importance of the quality of data in preference optimization.

\paragraph{IDK Ratio} Varying the ratio of IDK-labeled data reveals a nuanced and interesting trade-off. Higher ratios (0.1-0.7) intuitively improve AbQ F1 as the model learns to master the ability to abstain. However, too much IDK chosen data can lead to overly abstention resulting in decrease in overall abstain quality. Answer quality increases in sync with abstain quality showing an interesting balance. As the IDK ratio increases, the quality of correct responses does not decline significantly compared to the sharp rise in the model's refusal to answer. While the recall decreases as a result of fewer correctly answered questions, this way improves the precision of correct responses, ultimately enhancing the overall F1. However, when the model begins to overuse IDK (e.g., at extremely high ratio), this strategy ceases to work, as excessive abstention undermines correct answer coverage and utility. In addition, both DR and CUR scores consistently decrease as the IDK ratio increases, primarily due to the reduction in the proportion of \bluecheck\greencross\ and \bluecross\greencheck\ training data. The results suggest that moderate IDK ratios strike an optimal balance between precision and robustness, while aggressive reliance on IDK triggers diminishing returns.

\section{Comparison with SFT-Enhanced Baselines}
\label{appendix:sft_comparison}

To ensure fair comparison and address concerns about the SFT loss usage in our method, we conducted additional experiments where all baseline methods were evaluated on models that underwent SFT training using the same dataset as DTA. For P(True) and Logits baselines, we modified the SFT data to use (query, answer) pairs instead of (query, chosen) pairs to avoid performance degradation caused by "I don't know" patterns.

Notably, constructing the SFT dataset requires knowledge quadrant annotations for each sample, which are derived from the "Divide" stage of our DTA pipeline. Therefore, the SFT and ICL+SFT baselines benefit from a key contribution of our methodology.

\begin{table*}[t]
\centering
\begin{tabular}{l|cccc|ccc}
\toprule
\textbf{Model} & \textbf{Acc} & \textbf{Rec} & \textbf{Prec} & \textbf{F1} & \textbf{ARec} & \textbf{APrec} & \textbf{AF1} \\
\midrule
Original & 42.2 & 64.1 & 42.2 & 50.9 & 0.00 & 0.00 & 0.00 \\
RAAT & 46.2 & 70.2 & 46.2 & 55.7 & 0.00 & 0.00 & 0.00 \\
+ P(True) & 45.0 & 65.0 & 46.0 & 53.8 & 6.71 & 32.1 & 11.0 \\
+ Logits & 49.2 & 58.8 & 50.5 & 54.3 & 30.9 & 45.1 & 36.6 \\
+ Consistency & 51.4 & 69.0 & 50.7 & 58.5 & 16.3 & 58.4 & 25.4 \\
+ ICL & 46.8 & 71.2 & 46.8 & 56.5 & 0.00 & 0.00 & 0.00 \\
+ SFT & 52.2 & 37.4 & 69.1 & 48.5 & 80.7 & 42.9 & 56.0 \\
+ SFT \& P(True) & 48.1 & 69.5 & 48.9 & 57.4 & 6.8 & 35.7 & 11.4 \\
+ SFT \& Logits & 51.5 & 72.6 & 51.0 & 59.9 & 10.8 & 58.1 & 18.2 \\
+ SFT \& Consistency & 51.2 & 73.5 & 50.7 & 60.0 & 8.3 & 62.5 & 14.6 \\
+ SFT \& ICL & 59.7 & 63.1 & 58.1 & 60.5 & 53.1 & 63.6 & 57.9 \\
+ DTA & 64.1 & 63.7 & 65.5 & 64.6 & 65.0 & 61.7 & 63.3 \\
\bottomrule
\end{tabular}
\caption{Performance comparison with SFT-enhanced baselines on combined NQ, TriviaQA, and WebQ datasets. Results demonstrate that DTA significantly outperforms baseline methods even when they benefit from SFT training.}
\label{tab:sft_comparison}
\end{table*}

The results demonstrate that our DTA method consistently outperforms all baseline approaches, even when they benefit from SFT training. This validates the effectiveness of our approach beyond the training paradigm differences.





\section{Human Validation of GPT-4o Assessments}
\label{appendix:human_validation}

\begin{table}[h]
  \centering
  \small
  \begin{tabular}{l|c}
  \toprule
  \textbf{Method} & \textbf{Agreement with Human (\%)} \\
  \midrule
  GPT-4o Assessment & 93.0 \\
  Answer Matching & 76.0 \\
  \bottomrule
  \end{tabular}
  \caption{Human-AI agreement comparison. GPT-4o significantly outperforms traditional answer matching methods in determining retrieval knowledge boundaries.}
  \label{tab:human_agreement}
  \end{table}

To validate the reliability of GPT-4o as a judge for determining retrieval knowledge boundaries, we conducted human annotation experiments. We randomly sampled 100 (query, retrieval, answer) triples and had three human annotators independently label whether the retrieved passage contained the necessary information to answer the question. Final ground truth was established through majority voting.

\textbf{GPT-4o Evaluation:} "The context mentions the introduction of Bahamian dollar notes by the government in 1966, which directly implies that the Bahamian dollar is the kind of money to take to the Bahamas."

\textbf{Human Evaluation:} The context does not explicitly state that the Bahamian dollar is the currency of the Bahamas, making the inference less direct than GPT-4o suggests.

This case illustrates the nuanced differences in reasoning between human annotators and GPT-4o, where GPT-4o may make stronger inferences from contextual clues while humans prefer more explicit statements.

\section{Domain-Specific Evaluation}
\label{appendix:domain_specific}

To address concerns about generalizability to specialized domains, we conducted experiments on PubMedQA, a biomedical QA dataset. The knowledge boundary construction followed the same approach as our main experiments.

\begin{table*}[h]
\centering
\begin{tabular}{l|cccc|ccc}
\toprule
\textbf{Model} & \textbf{Acc} & \textbf{Rec} & \textbf{Prec} & \textbf{F1} & \textbf{ARec} & \textbf{APrec} & \textbf{AF1} \\
\midrule
Llama-2-7B & 50.7 & 78.7 & 50.7 & 61.6 & 0.0 & 0.0 & 0.0 \\
RAAT & 46.8 & 72.6 & 46.8 & 56.9 & 0.0 & 0.0 & 0.0 \\
+ P(True) & 46.7 & 63.5 & 48.1 & 54.7 & 16.3 & 38.9 & 22.9 \\
+ Logits & 45.1 & 56.2 & 44.3 & 49.5 & 25.0 & 48.6 & 33.0 \\
+ Consistency & 48.8 & 68.9 & 48.5 & 56.9 & 12.3 & 52.2 & 19.9 \\
+ ICL & 47.1 & 73.1 & 47.1 & 57.2 & 0.0 & 0.0 & 0.0 \\
+ DTA & 56.6 & 59.1 & 56.2 & 57.6 & 52.1 & 57.5 & 54.5 \\
\bottomrule
\end{tabular}
\caption{Performance on PubMedQA biomedical dataset. Despite distribution shift, DTA maintains strong performance and enables effective abstention compared to RAAT baseline.}
\label{tab:pubmedqa}
\end{table*}

The results show that while distribution shift affects performance, our DTA method still demonstrates strong capabilities in specialized domains, enabling appropriate abstention while maintaining overall accuracy improvements.

\section{Counterfactual Context Evaluation}
\label{appendix:counterfactual}

We evaluated our approach on ConFiQA-QA dataset to test robustness against counterfactual contexts. In this setup, counterfactual contexts are treated as noisy and original contexts as golden. We sampled 4,500 data points for alignment and reserved 1,500 for testing.

\begin{table*}[h]
\centering
\begin{tabular}{l|cccc|ccc}
\toprule
\textbf{Model} & \textbf{Acc} & \textbf{Rec} & \textbf{Prec} & \textbf{F1} & \textbf{ARec} & \textbf{APrec} & \textbf{AF1} \\
\midrule
Llama-2-7B & 41.4 & 75.8 & 41.4 & 53.5 & 0.0 & 0.0 & 0.0 \\
RAAT & 43.5 & 79.5 & 43.5 & 56.2 & 0.0 & 0.0 & 0.0 \\
+ P(True) & 41.5 & 63.7 & 41.1 & 49.9 & 14.6 & 43.3 & 21.8 \\
+ Logits & 47.9 & 74.3 & 46.2 & 56.9 & 16.0 & 60.5 & 25.3 \\
+ Consistency & 42.1 & 75.0 & 41.8 & 53.6 & 2.53 & 57.5 & 4.86 \\
+ ICL & 44.6 & 80.1 & 44.1 & 56.8 & 1.76 & 100.0 & 3.45 \\
+ DTA & 81.2 & 84.6 & 78.1 & 81.2 & 77.0 & 85.6 & 81.1 \\
\bottomrule
\end{tabular}
\caption{Performance on ConFiQA-QA dataset with counterfactual contexts. DTA achieves exceptional performance, demonstrating robustness against malicious attacks on RAG systems.}
\label{tab:confiqa}
\end{table*}

The results demonstrate exceptional performance on counterfactual contexts, with AF1 score exceeding 81.1\%, indicating that our method is robust against malicious attacks where counterfactual passages might be injected into RAG knowledge bases.

\section{Multi-hop Question Answering}
\label{appendix:multihop}

To evaluate performance on more complex reasoning tasks, we conducted experiments on HotpotQA, a multi-hop QA dataset. We derived training samples from the hard-level subset using chain-of-thought (CoT) prompting to establish model knowledge boundaries.

\begin{table*}[h]
\centering
\begin{tabular}{l|cccc|ccc}
\toprule
\textbf{Model} & \textbf{Acc} & \textbf{Rec} & \textbf{Prec} & \textbf{F1} & \textbf{ARec} & \textbf{APrec} & \textbf{AF1} \\
\midrule
Llama-2-7B & 27.1 & 44.9 & 27.1 & 33.8 & 0 & 0 & 0 \\
RAAT & 26.7 & 44.3 & 26.7 & 33.3 & 0 & 0 & 0 \\
+ P(True) & 27.2 & 39.0 & 26.4 & 31.5 & 9.2 & 33.0 & 14.4 \\
+ Logits & 33.7 & 39.8 & 29.9 & 34.1 & 24.4 & 49.2 & 32.6 \\
+ Consistency & 32.0 & 41.5 & 28.9 & 34.1 & 17.6 & 52.1 & 26.3 \\
+ ICL & 27.3 & 45.2 & 27.3 & 34.1 & 0 & 0 & 0 \\
+ DTA (trained on NQ, TriviaQA, WebQ) & 48.8 & 30.4 & 42.0 & 35.3 & 76.7 & 54.0 & 63.4 \\
+ DTA (trained on HotpotQA) & 59.8 & 52.0 & 49.7 & 50.9 & 71.5 & 76.9 & 74.1 \\
\bottomrule
\end{tabular}
\caption{Performance on HotpotQA multi-hop QA dataset. DTA demonstrates strong generalization ability and can appropriately abstain even with multiple-passage retrieval contexts.}
\label{tab:hotpotqa}
\end{table*}

Results show that even when retrieval knowledge comprises multiple passages, our method can still appropriately abstain from answering and demonstrates strong generalization ability across different training configurations.





\section{Prompts}
\label{appendix:prompts}
\subsection{Context Evaluation Prompt}
The following prompt is used to evaluate whether a context contains or implies the correct answer to a query:

\begin{tcolorbox}[colback=gray!10,boxrule=0.5pt]
  You are an expert at evaluating whether a context contains
the correct answer to a question. You should:

1. Check if the given answer can be found or directly
   implied by the context

2. Return a score of 1 if the context contains or directly
   implies the answer

3. Return a score of 0 if the context does not contain or
   support the answer

4. Provide a brief explanation for your decision

Respond in the following JSON format:

\{

    "score": 0 or 1,

    "explanation": "your explanation here"

\}
\end{tcolorbox}

\section{Implementation Details}
\label{sec:details}

\subsection{Our Method Implementation}
For our proposed approach, we train the model for 3 epochs using a cosine learning rate scheduler with an initial learning rate of 5e-5 and a warmup ratio of 0.1. The $\beta$ and $\gamma$ are set to 1.0 and 0.5 respectively for all experiments. The training process employs the Paged AdamW optimizer with 32-bit precision and a weight decay of 0.05. To balance computational efficiency and memory constraints, we set the batch size to 16 per device with 2 gradient accumulation steps, allowing for effective training on larger datasets while maintaining memory efficiency. The threshold $\delta$ used for $\KBparam$ to sample $N(=10)$ responses is 1.0. Moreover, experiments are conducted on NVIDIA A100 GPUs with 80G of memory. Fixed random seed of 0 is used and the experimental results are reported within a single run.  The versions of the libraries used in this work are as follows: accelerate version 0.34.2, transformers version 4.46.3, trl version 0.12.1 and vllm version 0.6.1.post2. And the dpo training process costs approximately 6 GPU hours.

\subsection{Baselines Implementation}
\label{appendix:baselines_implementation}
We implement several baseline detection methods for comparison:
\begin{itemize}
    \item \textbf{P(True)}: Following \citet{kadavath2022language}, we prompt the LLM to evaluate the correctness of its own answer. The prompt presents the original question and the model's proposed answer, asking for a binary True/False classification. We experiment with multiple confidence thresholds (0.3, 0.5, 0.7, 0.9) to determine the optimal cutoff for each experimental setting.
    \begin{tcolorbox}[colback=gray!10,boxrule=0.5pt]
      Question: [Question]

      Proposed Answer: [Predictions]

      Is the proposed answer:

      (A) True

      (B) False

      The proposed answer is:
      \end{tcolorbox}

      \item \textbf{Logits}: We implement three baselines using different logprob statistics of the output tokens: minimum (Min), mean (Mean), and last token (Last). The Min baseline, which uses the minimum logprob across all output tokens, is the only one reported in the paper as the other two approaches proved ineffective at enabling model abstention. We experiment with multiple logtis thresholds (-2.0, -1.0, 0.0) to determine the optimal cutoff for each experimental setting.
   
      \item \textbf{Self-Consistency}: We generate multiple responses (n=10) for each question and measure consistency among the generated answers. The system proceeds with answering if the most frequent response receives more than 5 votes; otherwise, it abstains. This approach helps identify cases where the model exhibits high uncertainty through response variability.

    \item \textbf{ICL}: We implement in-context learning using a prompt template with three carefully curated demonstration examples: one showing appropriate abstention for out-of-knowledge-boundary queries, and two showcasing correct answer generation for in-boundary queries. This balanced demonstration set helps the model learn both when to answer and when to abstain.
\end{itemize}

\section{Extended Related Work Discussion}
\label{appendix:extended_related}

Based on reviewer feedback, we provide additional discussion of relevant literature that complements our main related work section.

\subsection{Context-Faithfulness and Factuality Enhancement}

The knowledge boundary of Retrieval-Augmented Generation (RAG) is intrinsically linked to context-faithfulness \cite{zhou-etal-2023-context,bi2024context,bi2025parameters}. RAG extends a model's knowledge by incorporating external documents, which fundamentally requires the model to be faithful to the provided contextual information. Consequently, accurately perceiving these dynamic knowledge boundaries—the effective scope of a model's knowledge within a given context—is crucial. Research has explored leveraging the internal states of LLMs to enhance this perception of knowledge boundaries~\cite{ni2025towards}.  However, a core challenge arises when the model's parametric knowledge conflicts with the retrieved context, necessitating a balance in determining which knowledge source to prioritize.  To address this, strategies for fine-grained control over the model's reliance on parametric versus contextual knowledge have been proposed ~\cite{bi2025parameters}.  Concurrently, to improve adherence to context in RAG scenarios, alignment techniques such as Context-DPO \cite{bi2024context} have been developed to bolster context-faithfulness, particularly when knowledge conflicts occur.  A complicating factor is that efforts to enhance the factual accuracy of a model's internal knowledge can sometimes inadvertently degrade context-faithfulness, causing the model to over-rely on its parametric knowledge and disregard external inputs~\cite{bi2024factuality}.  In this light, enhancing reasoning capabilities through methods like in-context learning \cite{ge2025innate} may help models more effectively navigate the complex interplay between parametric knowledge and contextual information

\subsection{Uncertainty Expression and Knowledge Boundary Perception}

Our work is also related to research on verbalized confidence, where models express uncertainty through natural language rather than probability scores:  \citet{lin2022teaching} explores methods for teaching models to verbalize their confidence levels, which is conceptually related to our approach of teaching models to say "I don't know."  Research on when LLMs need retrieval augmentation \cite{ni2024llms} investigates mitigating overconfidence, which aligns with our goal of appropriate abstention in RAG systems. \citet{azaria2023internal} examines whether LLMs' internal states reveal when they are "lying", which provides insights into knowledge boundary detection that complement our external evaluation approach. \citet{ni2025towards} explores how to fully exploit LLM internal states to enhance knowledge boundary perception.

\section{Licensing}
Llama2-7B and Llama2-13B are released under the Meta Llama 2 Community License Agreement. Llama3-8B is released under the Meta Llama 3 Community License Agreement. All of them are accessible for academic usage and consistent with their intended use.

And three open-domain QA datasets, Natural Questions (NQ), TriviaQA, and WebQuestions (WebQ) are publicly available for academic research purposes, which is also consistent with their intended use.

\end{document}